\documentclass[10pt,journal,compsoc]{IEEEtran}
\usepackage{amsmath}
\usepackage{amssymb}
\usepackage{array}
\usepackage{graphicx}
\usepackage[usenames, dvipsnames]{color}
\usepackage{rotating}
\usepackage{multirow}
\usepackage{pgfplots}
\usepackage{algorithm}
\usepackage{algorithmicx}
\usepackage[noend]{algpseudocode}
\usepackage{hyperref}
\usepackage[margin=4pt,font=footnotesize,labelfont=bf,labelsep=endash,tableposition=top]{caption}
\usepackage{amsmath}
\usepackage{cases}
\usepackage{times}
\usepackage{subcaption}
\usepackage{epsfig}
\usepackage{graphicx}
\usepackage{amsmath}
\usepackage{amssymb}
\usepackage{multirow}
\usepackage{multicol}
\usepackage{algorithmicx}
\usepackage{enumitem}
\usepackage{arydshln}
\usepackage{bm}
\usepackage{enumitem}  

\newcommand{\ie}{\emph{i.e. }}
\newcommand{\eg}{\emph{e.g. }}

\pgfplotsset{compat=1.5}

\makeatletter
\newcommand{\printfnsymbol}[1]{
  \textsuperscript{\@fnsymbol{#1}}
}
\makeatother

\ifCLASSOPTIONcompsoc
  \usepackage[nocompress]{cite}
\else
  \usepackage{cite}
\fi

\ifCLASSINFOpdf
\else
\fi

\begin{document}

\title{PolarMask++: Enhanced Polar Representation for Single-Shot Instance Segmentation and Beyond\\ 
}

\author{
Enze Xie,
Wenhai Wang,
Mingyu Ding,
Ruimao Zhang,
Ping Luo
\IEEEcompsocitemizethanks{\IEEEcompsocthanksitem 
Enze Xie, Mingyu Ding and Ping Luo are with the Department of Computer Science, The University of Hong Kong. 
Wenhai Wang is with Nanjing University.
Ruimao Zhang is with the School of Data Science, The Chinese University of Hong Kong, Shenzhen and Shenzhen Research Institute of Big Data. 
Corresponding to Ping Luo (pluo@cs.hku.hk) and Enze Xie (xieenze@hku.hk). 
Enze Xie and Wenhai Wang contributed equally. 
}
}

\markboth{IEEE Transactions on Pattern Analysis and Machine Intelligence}%
{Shell \MakeLowercase{\textit{et al.}}: Bare Demo of IEEEtran.cls for Computer Society Journals}

\IEEEtitleabstractindextext{%

\begin{abstract}
Reducing the complexity of the pipeline of instance segmentation is crucial for real-world applications.
This work addresses this issue by introducing an anchor-box free and single-shot instance segmentation framework, termed PolarMask,
which reformulates the instance segmentation problem as predicting the contours of objects in the polar coordinate, with several appealing benefits.
(1) The polar representation unifies instance segmentation (masks) and object detection (bounding boxes) into a single framework, reducing the design and computational complexity.
(2) Two modules are carefully designed~(\ie soft polar centerness and polar IoU loss)  to sample high-quality center examples and optimize polar contour regression, making the performance of PolarMask does not depend on the bounding box prediction results and thus becomes more efficient in training.
(3) PolarMask is fully convolutional and can be easily embedded into most off-the-shelf detection methods. To further improve the accuracy of the framework, a Refined Feature Pyramid is introduced to further improve the feature representation at different scales, termed PolarMask++.
Extensive experiments demonstrate the effectiveness of both PolarMask and PolarMask++, which achieve competitive results on instance segmentation in the challenging COCO dataset with single-model and single-scale training and testing, as well as new state-of-the-art results on rotate text detection and cell segmentation.
We hope the proposed polar representation can provide a new perspective for designing algorithms to solve single-shot instance segmentation. 
The codes and models are available at:
\href{https://github.com/xieenze/PolarMask}{{\tt github.com/xieenze/PolarMask}}.

\end{abstract}

}

\maketitle

\IEEEdisplaynontitleabstractindextext

\IEEEpeerreviewmaketitle

\section{Introduction}\label{sec:int}

\begin{figure}[t]
\centering
\includegraphics[width=0.45\textwidth]{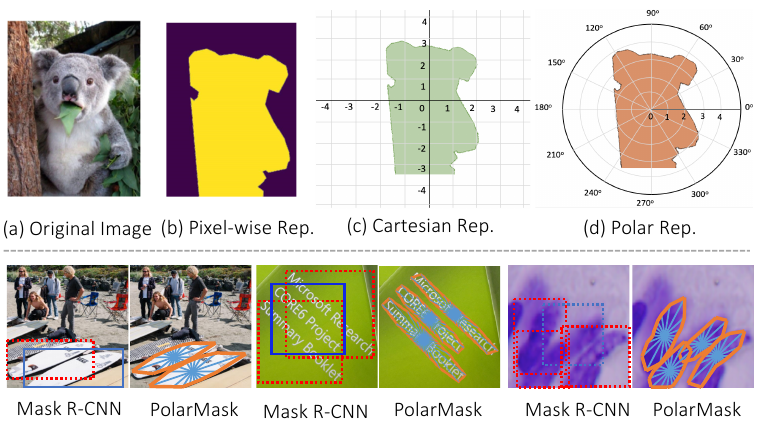}
\caption{
\textbf{First Row: Comparisons between Cartesian representation and polar representation.} ``Rep.'' indicates representation.
(a) shows an original image. (b) is its corresponding pixel-wise mask. (c) and (d) represent the mask by using contours in the Cartesian coordinates and Polar coordinates (a center and ray lengths at twelve angles), respectively.
\textbf{Second Row: Comparisons between our PolarMask and the popular Mask R-CNN,} where  PolarMask is a bounding box-free method because polar coordinate is a unified and more flexible representation than bounding box (\eg a bounding box is a polar mask with only four rays). %
Polar representation has more advantages than bounding boxes. For example, when the object instances are rotated and crowded, the original bounding boxes are not tied and the adjacent boxes are easily suppressed by non-maximum suppression (NMS) as shown by the boxes in red dots, leading to mis-detected instances. Our PolarMask system directly predicts center point and ray lengths of the object in the polar coordinate without relying on bounding box, thus being suitable in challenging situations such as heavy rotations and crowded objects.
}
\label{fig:intro}
\end{figure}

\IEEEPARstart{I}{nstance} segmentation is one of the most fundamental tasks in computer vision.
As the object mask of an object instance provides more accurate boundary information than its bounding box does, instance segmentation improves the performances of numerous downstream vision applications, such as text detection and recognition in visual navigation, cell segmentation in biotechnology, defect localization in manufacturing, so on and so forth.

However, instance segmentation is challenging because it requires predicting both the location and the semantic mask of each object instance in an image.
Therefore, instance segmentation has been typically solved by firstly detecting bounding boxes and secondly performing semantic segmentation within each detected bounding box. This is a two-stage pipeline adopted by two-stage instance segmentation methods, such as Mask R-CNN~\cite{maskrcnn}, PANet~\cite{panet} and Mask Scoring R-CNN~\cite{msrcnn}. 
The above pipeline is straightforward and has achieved good performance, but may suffer from heavy computational overhead, that its their ability in real-time applications in limited.

To address the above issue, recent trend has spent more effort in designing simpler single-stage pipeline for bounding box detection~\cite{densebox,focalloss,FCOS,reppoints,objectspoints,centernet,foveabox} and   instance segmentation~\cite{yolact,tensormask,extremenet,deepsnake}.
This is also the main focus of this work,
\textit{
    which designs a conceptually simple and unified mask prediction module that can be easily plugged into many off-the-shelf detectors, enabling instance segmentation and rotated object detection.
}

Intuitively, instance segmentation is usually solved by binary classification in a spatial layout surrounded by bounding boxes, as shown in Figure~\ref{fig:intro}(b). 
Such pixel-to-pixel correspondence prediction is luxurious, especially in the single-shot methods. 
Instead, we show that masks can be recovered successfully and effectively if the contour is obtained. 
A representative approach to predict contours is shown in Figure~\ref{fig:intro}(c), which localizes the Cartesian coordinates of the points composing the contour. We term it ``cartesian representation''. 
An alternative is ``polar representation'', which applies the angle and the distance as the coordinate to localize points, as shown in Figure~\ref{fig:intro}(d).
The polar representation with several advantages, is suitable for instance segmentation.
Firstly, the origin point of the polar coordinate can be regarded as the center of an object. 
Secondly, starting from the origin, a point on the contour can be simply determined by its distance and angle with respect to the origin.
Thirdly, the angle is naturally directional and thus renders it convenient to connect multiple points into a contour.

The cartesian representation may exhibit the above first two properties, but it lacks the advantage of the third one. For example, polar representation can naturally obtain the order of a sequence of points for an object according to a set of different angles. For each point, we only need to regress one parameter (the length of the ray, that is the distance between the center point and the contour). However, for cartesian representation, we cannot naturally give a clear order of the points in the cartesian coordinate without angles. Moreover, for each point in cartesian representation, we need to regress two parameters,  $x$ and $y$ coordinates.
As a result, cartesian representation is only suitable to represent the bounding box, while polar representation is a unified representation and can represent both bounding box and contour of object because the bounding box is just the simplest contour with 4 angles and rays.
As shown in Figure~\ref{fig:intro}'s column 2, polar representation can enable box-free instance segmentation, which has large advantages over box-based instance segmentation methods, \textit{e.g.} Mask R-CNN, on crowd scene. For example, polar representation has shown great advantages over cartesian representation on remote sensing Object detection, scene text detection and cell instance segmentation\cite{polardet,sapr,zhou2020objects,textray}.

The polar representation enables us to reformulate instance segmentation as instance center classification and dense coordinate regression in the polar coordinate, whereby
we propose PolarMask, an anchor-box-free and single-shot instance segmentation method. 
Specifically, PolarMask takes an image as input and predicts the distance from a sampled positive location (\ie a candidate object's center) with respect to the object's contour from each angle, and then assembles the predicted points to produce the final mask. 
To leverage the benefits of the polar representation, 
we introduce two novel modules, \textit{i.e.}, the soft polar centerness and the polar IoU loss, aiming sample high-quality center examples and ease optimization of the dense coordinate regression problem.

In summary, the proposed PolarMask has several appealing benefits compared to prior arts. 
(1) The polar representation unifies instance segmentation and detection into a single framework, integrating the design of the above two technical routes been unified while reducing computational complexity. 
(2) The performance of PolarMask does not depend on the results of bounding box predictions. By removing the bounding-box branches~(which is indispensable for Mask R-CNN), PolarMask has a lower computational cost, being more efficient in training.
(3) It is fully convolutional and can be easily embedded into most off-the-shelf detection systems.
For example, we instantiate the proposed method by embedding it into the recent object detector FCOS~\cite{FCOS}, which is a simple and low-cost pipeline.  It is also worthy noting that PolarMask can be also used with other detectors such as RetinaNet~\cite{focalloss} and YOLO~\cite{yolo}.

\begin{figure*}[!t]
\begin{center}
\includegraphics[width=0.99\textwidth]{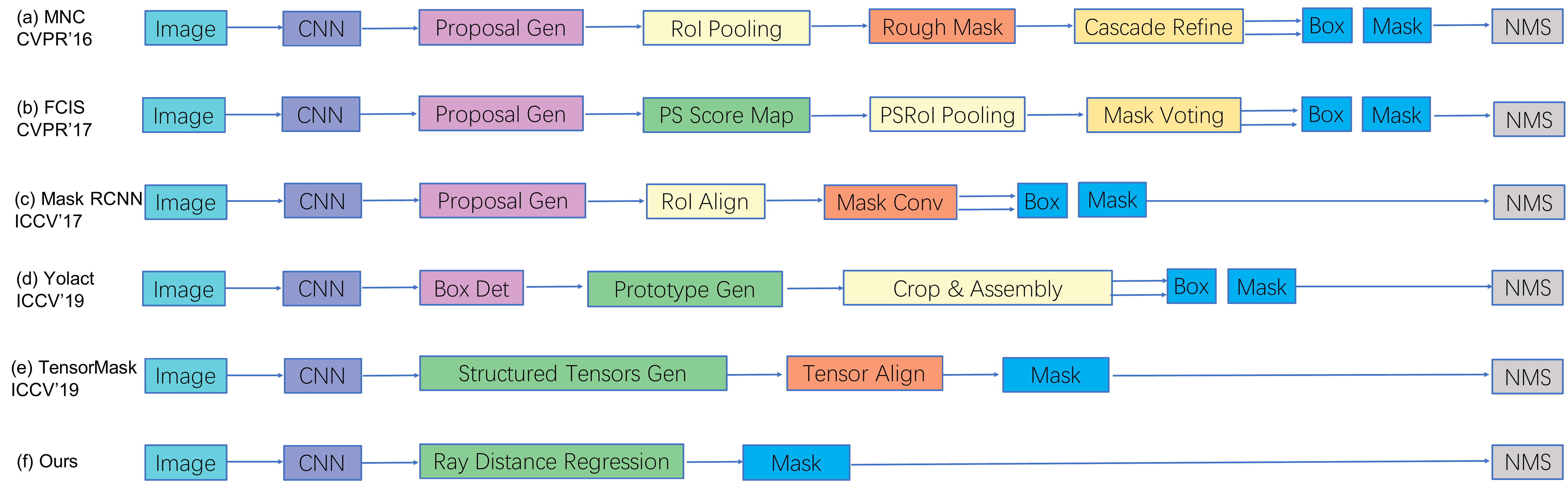}
\caption{\textbf{The overall pipeline of PolarMask compared to the previous representative methods.} ``Gen'' means generation. ``Det'' represents detection and ``Conv'' is convolution operation. 
We can see all the two-stage methods (\textit{e.g.} MNC(a), FCIS(b) and Mask R-CNN(c)) and one stage method Yolact(d) rely on box detection results, following the paradigm of ``detect then segment''.
Although TensorMask(e) does not need box predictions, its architecture is also complex. It models instances as 4D tensor by using ``Structured Tensors Generation'' and uses ``Tensor Align'' to align features, leading to slow inference speed.
The proposed PolarMask(f) is much simpler than other methods in pipeline design and does not need box predictions.} 
\label{fig:methods}
\end{center}
\vspace{-3mm}
\end{figure*}

As shown in Figure~\ref{fig:methods}(f), our resulting pipeline is as simple and clean as FCOS compared to other existing works in (a-e). PolarMask introduces {negligible} computational overhead, with both simplicity and efficiency, which are the two key factors for single-shot instance segmentation.
In this example, PolarMask actually generalizes FCOS in polar representation by optimizing polar centerness and polar IoU. 
To further improve its accuracy, a Refined Feature Pyramid module is proposed to further improve the feature fusion ability at different scales.
As a result, our framework takes advantage of the polar representation, which is much simpler and has fewer modules and efficient processes than the ones based on bounding box prediction as shown in Figure \ref{fig:methods}.
Without bells and whistles, PolarMask relatively improves the mask accuracy by about 25\%, showing considerable gains under strict localization metrics. 
For instance, with only single-scale testing, it achieves competitive or state-of-the-art performances on multiple tasks including instance segmentation, rotated text detection and cell segmentation, such as 38.7\% mask mAP on COCO \cite{coco}, 85.4\% F-measure on ICDAR2015 \cite{2015icdar}, and 74.2\% mAP on DSB2018 \cite{dsb}.

The main \textbf{contributions} of this work are three-fold.
(1) 
We introduce a new perspective to design a single-shot instance segmentation framework, PolarMask, which predicts instance masks and rotated objects in the polar coordinate in an effective and efficient manner.
(2) With the polar representation, we propose the polar IoU loss and the soft polar centerness for instance center classification and dense coordinate regression.
We show that the proposed IoU loss in polar space can largely ease the optimization and improve accuracy, compared with the standard loss such as the smooth-\l1 loss. 
In parallel, soft polar centerness improves the previous centreness loss in FCOS \cite{FCOS} and  PolarMask~\cite{polarmask}, leading to further boost in performance.
(3)
Rich experiments show that state-of-the-art performances of object instance segmentation and rotated object detection can be achieved with low computational overhead in multiple challenging benchmarks.

\section{Related Work}
\subsection{General Object Detection}
We also review algorithms for general object detection, which could be categorized into two-stage and one-stage methods.

\textbf{Two-stage detectors} mostly follow the R-CNN~\cite{rcnn} pipeline, which firstly generates a set of object proposals and then refines the proposals by a subnetwork in each region. For example, SPPNet~\cite{sppnet} and Fast R-CNN~\cite{fastrcnn} have similar region-wise feature extractors, while Faster R-CNN~\cite{fasterrcnn} proposes a Region Proposal Network (RPN) to generate proposals. R-FCN~\cite{rfcn}
introduces a position-sensitive RoI Pooling technique to reduce computations of the region-wise subnetwork. These methods improve R-CNN's performance. Based on them,  HyperNet~\cite{hypernet} and FPN~\cite{fpn} involve multi-layer features,
accounting for objects in various
scales. 

\textbf{One-stage object detection} is another popular research topic. It aims to design a simple pipeline to reduce the cost of the two-stage methods, leading to efficient computation.
These approaches often drop the step of proposal generation by directly predicting the final outputs following the merit of the classic sliding window strategy.
For example, YOLO~\cite{yolo,yolov2} is a representative single-stage object detector. By using a sparse feature map for object detection, it achieved nearly real-time speed but scarifies accuracy. To improve the accuracy and tackle objects with multiple scales,
SSD~\cite{SSD} employs the feature maps that are produced by multiple layers and detects the object with different sizes in different layers. DSSD~\cite{dssd} extended SSD by utilizing a deconvolution module to fuse the low-level feature and the high-level feature, involving more contextual information for the low-level detector. A similar strategy is also proposed in FPN and RetinaNet~\cite{focalloss}. Meanwhile, RetinaNet further proposes focal loss to address the problem of foreground-background class imbalance. 

Recent trends focus on anchor free one-stage detectors~\cite{FCOS,reppoints,objectspoints,cornernet,foveabox}. For example, Cornetnet~\cite{cornernet} directly predicts the object heatmap of top-left and bottom-right and then uses embedding vectors to group them. FCOS~\cite{FCOS} abandons the anchor design by directly regressing four distances between the center and the bounding box. 
Reppoints~\cite{reppoints} uses deformable convolution to extract features around the boxes. However, none of the above works explored objects in polar space.

\subsection{Rotated Object Detection}

Rotated object detection is also related to this work.
It is a challenging task beyond traditional object detection with rich applications such as scene text detection in the real world. 
Recent advantages in scene text detection are based on deep learning.  For example, TextBoxes~\cite{textboxes} modifies anchors and kernels of SSD~\cite{SSD} to detect large-aspect-radio scene texts.  EAST~\cite{east} adopts FCN~\cite{fcn} to predict a text score map and a final bounding box in the text region.
RRD~\cite{rrd} extracts two types of features for classification and regression respectively for long text line detection.
Based on Faster R-CNN, RRPN~\cite{rrpn} adds rotation to both anchors and RoIPooling to detect multi-oriented text regions.
SPCNet~\cite{spcnet} uses Mask R-CNN to detect arbitrary-shape text and add text context module and re-score module to further suppress false positives.
PSENet~\cite{psenet}  segments the text kernels map and the entire text regions map, and then uses progressively scale expansion to reconstruct the whole text instances.
PAN~\cite{pan} is based on PSENet by using a light-weight backbone network and FPN. It also learns an embedding vector to distinguish which pixels belong to the corresponding text instance.

Instead of specially designed modules for rotated text detection, we present a unified polar representation to handle not only instance segmentation but also rotated object detection, by treating scene text as a special case of the mask. In this way, text detection can be easily integrated into PolarMask, and benefit from the advantages of polar representation, resulting in state-of-the-art performance.

\subsection{Polar Representation in Vision Applications}

Before the deep learning era, polar representation has been used in contour extraction and tracking.
For example, Denzler \textit{et al}~\cite{activeray} introduce \textit{active rays} to describe the contour extraction as an energy minimization problem. In practice, the authors use a reference point within the object's contour and shoot rays in different directions from the reference point.
They also introduce an energy term to describe the internal elasticity of the rays.
Their method can be applied to tracking a pedestrian approaching the camera. 
In the deep learning era, StarDist~\cite{celldet} firstly use polar representation to detect cells in microscopic images. They predict a \textit{star-convex polygon} for every pixel in the feature map. The first branch of StarDist predicts object probabilities on a binary mask to classify two categories~(\textit{i.e.} `cell' and `background'). The second branch predicts star-convex polygon distance, to calculate the Euclidean distance of pixel belonging to an object and the pixel on the object boundary by simply following each radial direction \textit{k}.
DARNet~\cite{darnet} use deep active rays for automatic building segmentation. They use polar coordinates to parameterize a polygon-based contour. This contour then evolves to minimize its energy via gradient descent. The whole training process of DARNet is end-to-end. 
More related work is ESE-Seg~\cite{ese_seg}, which is based on Yolo v3. The key idea of ESE-Seg is to calculate the inner center of object and sample contour points according to the angles at a fixed interval around the inner-center point, where the distance from center to the contour is named shape vector. 
They also shorten shape vectors and resist noise through Chebyshev polynomial fitting.
Finally, they add a regression branch on Yolo to train the network.
Following PolarMask, some recent articles~\cite{polardet,sapr,zhou2020objects,textray} also adopt polar representation in their research area, such as scene text detection and rotate object detection in remote sensing, which shows strong potential of polar representation for instance-level detection.

\section{Methodology}

Firstly, we reformulate instance mask segmentation in the polar coordinate space in section \ref{sec:overview}. Secondly, we introduce the polar centerness and polar IoU loss functions to optimize the polar coordinate regression problem in section \ref{sec:RepresentOptim}.
Thirdly, section \ref{sec:Architecture} provides details of the architecture of PolarMask++, where
a Refined Feature Pyramid module is proposed to improve the capacity to detect small objects.
The procedures of label generation and model optimization are presented at the end of this section.

\subsection{Overview of Mask Segmentation in Polar Coordinate \label{sec:overview}} 

\begin{figure}[!t]
\begin{center}
\includegraphics[width=0.49\textwidth]{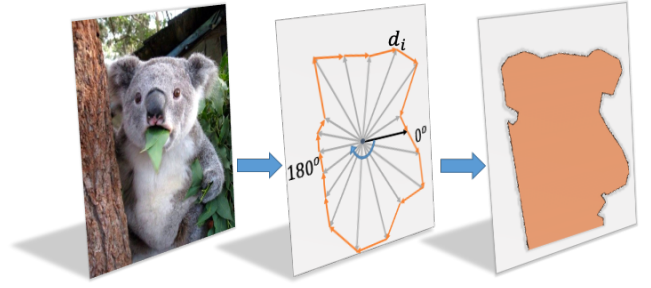}
\caption{\textbf{Mask Assembling}. Polar Representation provides a directional angle. The contour points are connected one by one start from $0^{\circ}$~(bold line) and assemble the whole contour. The mask is naturally obtained as the pixels inside the contour are the mask result.}
\label{fig:polarseg}
\end{center}
\vspace{-5mm}
\end{figure}

\textbf{Polar Representation.} 
As shown in Fig.\ref{fig:polarseg}, given a mask of an object instance, we sample a center point of the object, denoted as $(x_{c}, y_{c})$, and a set of points located on the contour of the object, denoted as $\{(x_{i}, y_{i})\}_{i=1}^N$. And then starting from the center, $n$ rays are emitted uniformly with the same angle interval $\Delta\theta$. For example, $n=36$ and $\Delta\theta=10^{\circ}$ represent 36 rays with $10^{\circ}$ between two adjacent rays.
The length of each ray is calculated as the distance between the center and the point on the contour. In this case, we could model a mask in the polar coordinate using one center and $n$ rays. Since the angle interval is a constant, only the lengths of the rays need to be learned. Therefore, we could reformulate  instance segmentation as instance center classification and regression of rays' lengths in the polar coordinate.

\textbf{Mass Center.} To represent an object's center, we consider both the bounding box's center and the mass center and evaluate the upper bound of the mask segmentation performance of them (details in Figure~\ref{fig:upper bound}).
We find that the mass center is more advantageous than the box center because the mass center has a larger probability of falling inside an instance compared to its box center. Although for some extreme cases such as a ``donut'', neither the mass center nor the box center lies inside the instance, the mass center is applicable to most of the cases better than the box center.

\textbf{Center Samples.} A location $(x, y)$ in an image is considered as a center (positive) sample if it falls into a certain range around the mass center of an object. Otherwise, it is treated as a negative sample. We define the range for sampling positive pixels to be 1.5 times the strides~\cite{FCOS} of the feature map from the mass center to the left, top, right, and bottom of the bounding box. 
As a result, 9$\sim $16 pixels around the mass center of each instance would be treated as positive examples, leading to two advantages. 
(1) This would increase the number of positive samples to avoid imbalance between the positive and negative samples in training. 
(2) More candidate points would represent the mass center of an instance more accurately.

\textbf{Distance Regression of Ray in Training.} 
Given a center point of a positive sample $(x_{c}, y_{c})$ and a set of points on its contour, the length of $n$ rays are denoted by $\{d_{1}, d_{2}, \ldots, d_{n}\}$ as shown in Figure \ref{fig:polarseg}. In this case, instance segmentation is treated as length regression of rays. 
The detailed computations are presented in section~\ref{labelgen}. Here we introduce two special cases.
(a) If a ray intersects with the contour in more than one point (\ie a ``concave'' boundary), the point with the maximum length with respect to the center is chosen to represent the ray.
(b) If a ray is connected to a center that lies outside the mask and it does not intersect with the contour given the angle $\Delta\theta$, we define its length as a small constant $\epsilon$, for example, $\epsilon=10^{-6}$.

\begin{figure}[t]
\begin{center}
\includegraphics[width=0.49\textwidth]{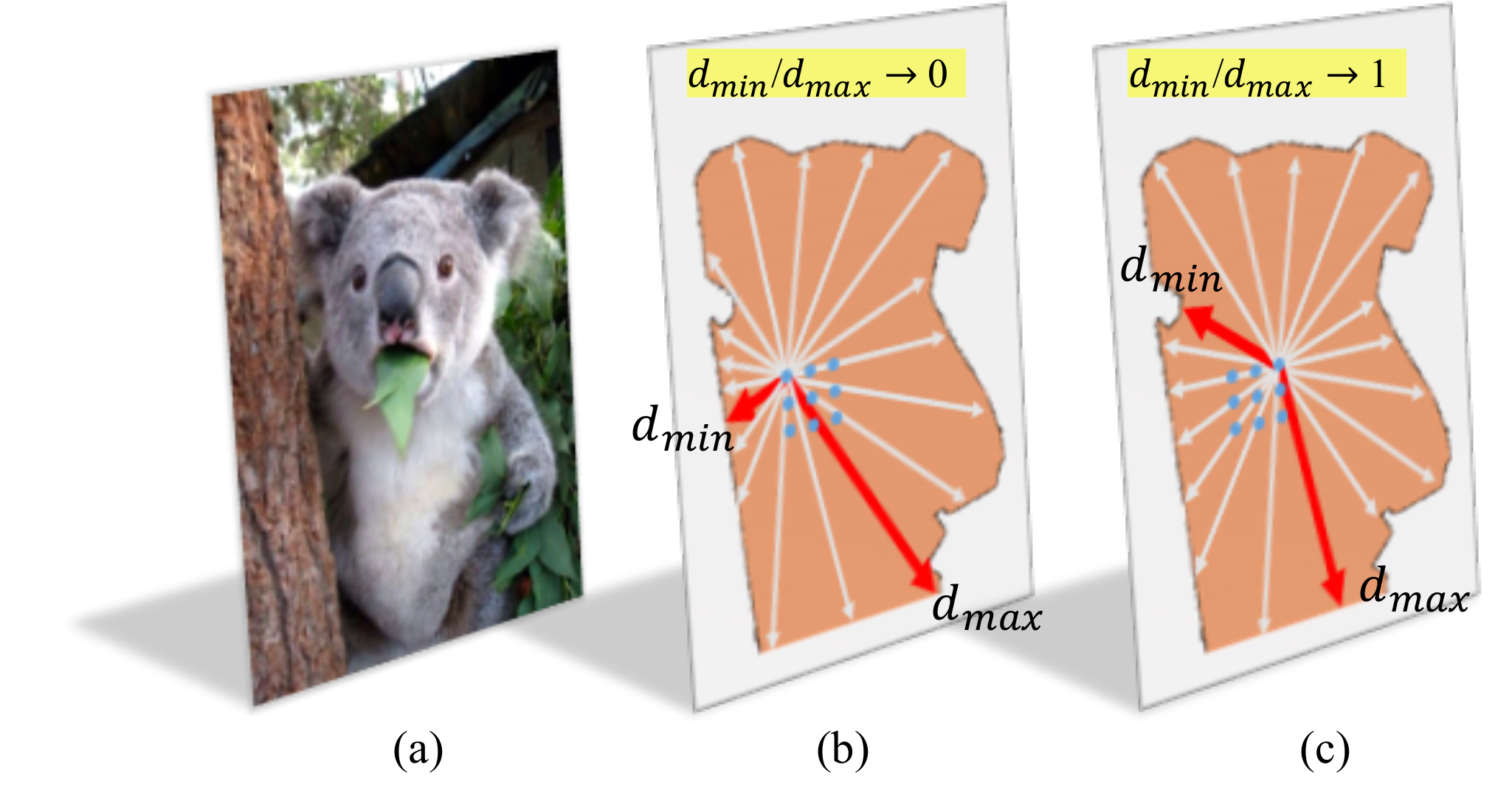}
\caption{\textbf{Polar Centerness}. 
(a) is the original image. 
(b) is the rays regression with a inappropriate center.
(c) is the rays regression with an appropriate center.
$d_{min}$ and $d_{max}$ are the minimum and maximum distance of rays.
Polar Centerness is used to down-weight such regression tasks as the high diversity of rays' lengths as shown in red lines in the middle plot. These examples are  hard to optimize and produce low-quality masks.
In (b), the $d_{min}/d_{max} \rightarrow 0$ 
and in (c), the $d_{min}/d_{max} \rightarrow 1$ , 
which means the positive sample in (c) is better than (b) and has higher loss weight during training.}
During inference, the polar centerness predicted by the network is multiplied to the classification score, thus can down-weight the low-quality masks.
\label{fig:ctness}
\end{center}
\vspace{-3mm}
\end{figure}

The training stage of length regression is non-trivial because of two reasons. Firstly, length regression is a dense distance regression task since every training sample has $n$ rays such as $n=36$. It may cause an imbalance between the regression loss and the center classification loss. Secondly, $n$ rays of each instance are correlated. They should be jointly trained, rather than being treated as a set of independent tasks. The above difficulties will be solved by the polar IoU loss function as discussed in Section~\ref{iouloss}.

\textbf{Mask Assembling in Testing.} 
After training, the network produces the confidence scores of the center and the rays' lengths.
We assemble masks from at most 1,000 top-scoring predictions per feature level in a Feature  Pyramid  Network (FPN).
The top predictions from all levels are merged and non-maximum suppression (NMS) is applied to yield the mask. 
Here we introduce the mask assembling and the  NMS procedures.

Given a center location $(x_{c}, y_{c})$ and the lengths of $n$ rays $\{d_{1}, d_{2}, \ldots, d_{n}\}$, we can calculate the position of each corresponding contour point $(x_{i},y_i)$,
\begin{equation}
 x_{i} = \cos\theta_{i} \times d_i + x_{c}
\end{equation}
\begin{equation}
 y_{i} = \sin\theta_{i} \times d_i + y_{c}.
\end{equation}
As shown in Figure~\ref{fig:polarseg}, starting from $0^{\circ}$, the contour points are connected one by one and finally assembles the entire contour as well as the mask.

We apply NMS to remove redundant masks. To simplify the process, we calculate the smallest bounding boxes of masks and then apply NMS based on the IoU of the generated bounding boxes.

\subsection{Polar Representation Optimization\label{sec:RepresentOptim}}
This section describes how to optimize the prediction of centerness and the regression of rays.

\subsubsection{Polar Centerness Prediction\label{centerness}}
Centerness~\cite{FCOS} is introduced to suppress low-quality bounding boxes in object detection. However, simply applying centerness in the polar space is sub-optimal because it is designed for regular bounding boxes but not masks. 

Here we define polar centerness. Let $\{d_{1}, d_{2}, \ldots, d_{n}\}$ be a set of lengths of $n$ rays of an instance,
the polar centerness could be  represented by,
\begin{equation}
{ \rm Polar~Centerness } 
= \sqrt{\frac{\min(\{d_{1}, d_{2}, \ldots, d_{n}\})}
        {\max(\{d_{1}, d_{2}, \ldots, d_{n}\})}},
\end{equation}
which assigns higher centerness to a location  if the $\min(\{d_{1}, d_{2}, \ldots, d_{n}\})$ value  and the $\max(\{d_{1}, d_{2}, \ldots, d_{n}\})$ value are close.

The above equation could localize the object center efficiently, as shown in Figure~\ref{fig:ctness}. 
However, we find that sometimes the $\min(\{d_{1}, d_{2}, \ldots, d_{n}\})$ value of the best positive sample  is not close to the $\max(\{d_{1}, d_{2}, \ldots, d_{n}\})$ value, especially for complex shapes, 
making the value of polar centerness small and resulting in
two drawbacks. 
(1) The weight of positive samples becomes small and this is not optimal for solving ray regression. 
(2) In test, the final scores of objects would be also small because they rely on the centerness prediction.

We improve it by introducing a soft mechanism, termed as soft polar centerness,
to bridge the gap between $\min(\{d_{1}, d_{2}, \ldots, d_{n}\})$ and $\max(\{d_{1}, d_{2}, \ldots, d_{n}\})$. 
In details, we divide $d_i$ according to its angle into four subsets, 
\begin{equation}
\begin{aligned}
D_1&=\{d_{1}, \ldots, d_{\frac{n}{4}}\} \in [0^{\circ}, 90^{\circ}),\\
D_2&=\{d_{\frac{n}{4}+1}, \ldots, d_{\frac{n}{2}}\} \in [90^{\circ}, 180^{\circ}),\\
D_3&=\{d_{\frac{n}{2}+1}, \ldots, d_{\frac{3n}{4}}\} \in [180^{\circ}, 270^{\circ}),\\
D_4&=\{d_{\frac{3n}{4}+1}, \ldots, d_{n}\} \in [270^{\circ}, 360^{\circ}).
\end{aligned}
\end{equation}
Then the soft polar centerness is defined as,
\begin{equation}
{ \rm Soft~Polar~Centerness } 
= \sqrt{
        \frac{F(D_1)}
        {F(D_3)}
        \times
        \frac{F(D_2)}
        {F(D_4)}
        },
\end{equation}
where $F$ is a function to calculate a value given a set $D_i$. We investigate three different functions, including~(1) the mean of $D_i$, (2)~the maximum value of $D_i$, and
(3)~the first value of $D_i$. Through extensive experiments, we found that the results are comparable and all of them improve the performance of the original polar centerness.

\begin{figure*}[!t]
\begin{center}
\includegraphics[width=0.99\textwidth]{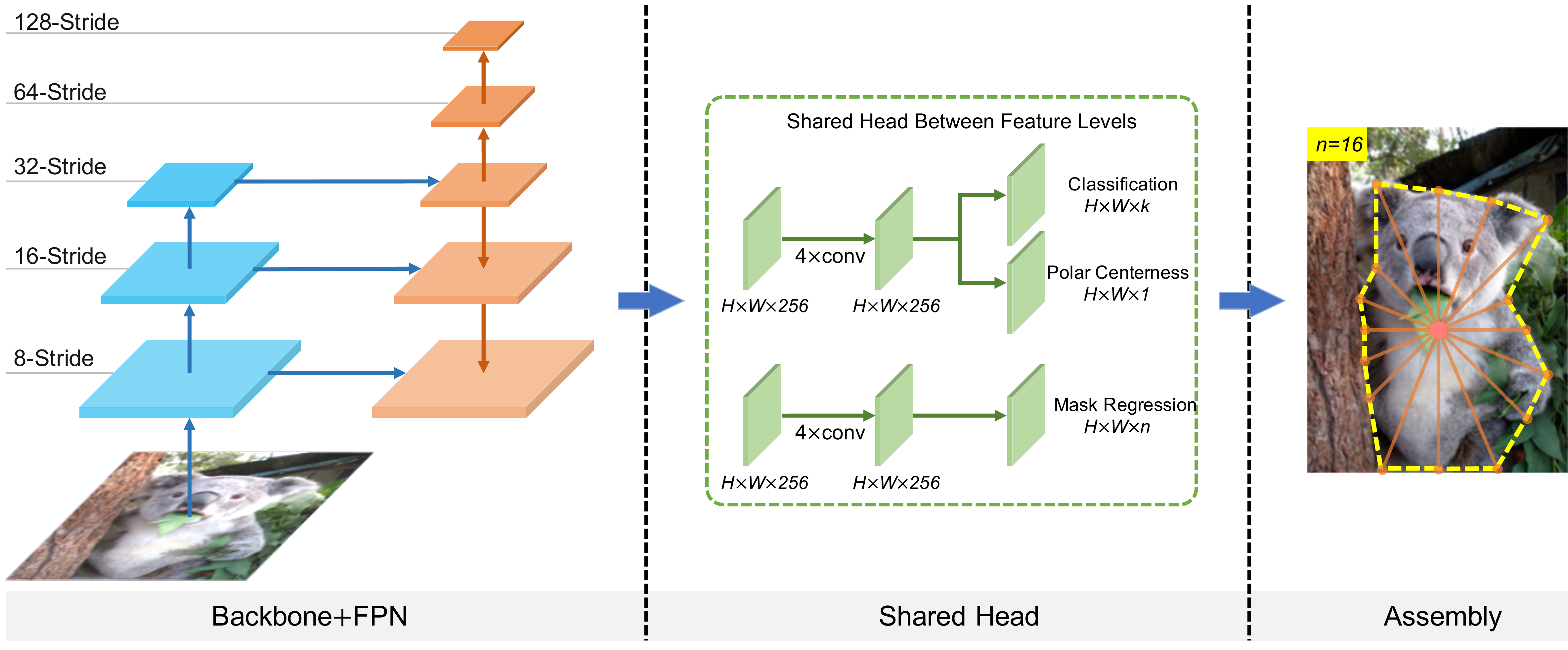}
\caption{\textbf{The overall pipeline of PolarMask system.} The left part contains the backbone and feature pyramid to extract features of different levels. The middle part is the two heads for classification and polar mask regression. $H, W, C$ are the height, width, channels of feature maps, respectively, and $k$ is the number of categories (e.g., $k =80$ on the
COCO dataset), $n$ is the number of rays (e.g., $n=36$).}
\label{fig:pipeline}
\end{center}
\vspace{-3mm}
\end{figure*}

In the implementation, as shown in Figure~\ref{fig:pipeline}, we add a branch with a single layer to predict the soft polar centerness, which is in parallel with the classification branch. 
The polar centerness predicted by the network is multiplied by the classification score, thus reducing the weight of low-quality masks.
Experiments show that soft polar centerness improves accuracy especially under strict localization metrics such as AP$_{75}$.

\subsubsection{Polar Ray Regression \label{iouloss}}
As discussed above, the method of polar segmentation converts the task of instance segmentation into a set of ray regression problems. 
In object detection and segmentation, the smooth-\l1 loss~\cite{rcnn} and the IoU loss~\cite{unitbox} are the two widely-used loss functions to solve the regression problems. 
However, both of these functions have certain drawbacks. First, the smooth-\l1 loss does not capture  correlations between samples of the same objects, thus resulting in less accurate localization. 
Second, although the IoU loss directly optimizes the metric of interest (IoU), 
computing the IoU of the predicted mask and its ground-truth is difficult to implement in parallel.

This work derives an easy and effective method to compute the mask IoU in the polar space and defines the Polar IoU loss function to optimize the model and achieves  competitive performance. 
Let $d^{\min}_i=\min(d_i,d^*_i)$ and $ d^{\max}_i=\max(d_i,d^*_i)$, where $d_i$ and $d^*_i$ indicate the target and the predicted length of the $i$-th ray respectively.
The Polar IoU  is computed as
\begin{equation}
  {\rm Polar~IoU}  = \frac{\sum_{i=1}^{n}d^{\min}_i}
        {\sum_{i=1}^{n}d^{\max}_i},
\label{PolarIoU}
\end{equation}
where we draw the connection between the ray regression and the polar IoU of the predicted mask.
Then we can define the Polar IoU loss function as the binary cross entropy (BCE) loss of the Polar IoU to optimize the length of each ray.
Since the optimal IoU is always $1$, the polar IoU loss function can be represented by the negative logarithm of the Polar IoU,
\begin{equation}
{  \rm Polar~IoU~Loss }
= \log \frac{\sum_{i=1}^{n}d^{\max}_i}
        {\sum_{i=1}^{n}d^{\min}_i}.
\label{IoULoss}
\end{equation}
The above polar IoU loss function exhibits two advantageous properties. (1) It is differentiable by using back-propagation and it is easy to implement parallel computations,  facilitating a fast training process. (2) It improves the overall performance by a large margin compared with the smooth-\l1 loss function by predicting all the regression rays as a whole, rather than treating them independently.  (3) Moreover, Polar IoU loss is able to automatically balance between the classification loss and the regression loss of dense distance prediction. We will discuss this in detail in experiments.

\subsubsection{Discussions of the Effectiveness of Polar IoU \label{EffectiveIoU}}

\begin{figure}[t]
\centering 
\includegraphics[width=0.49\textwidth]{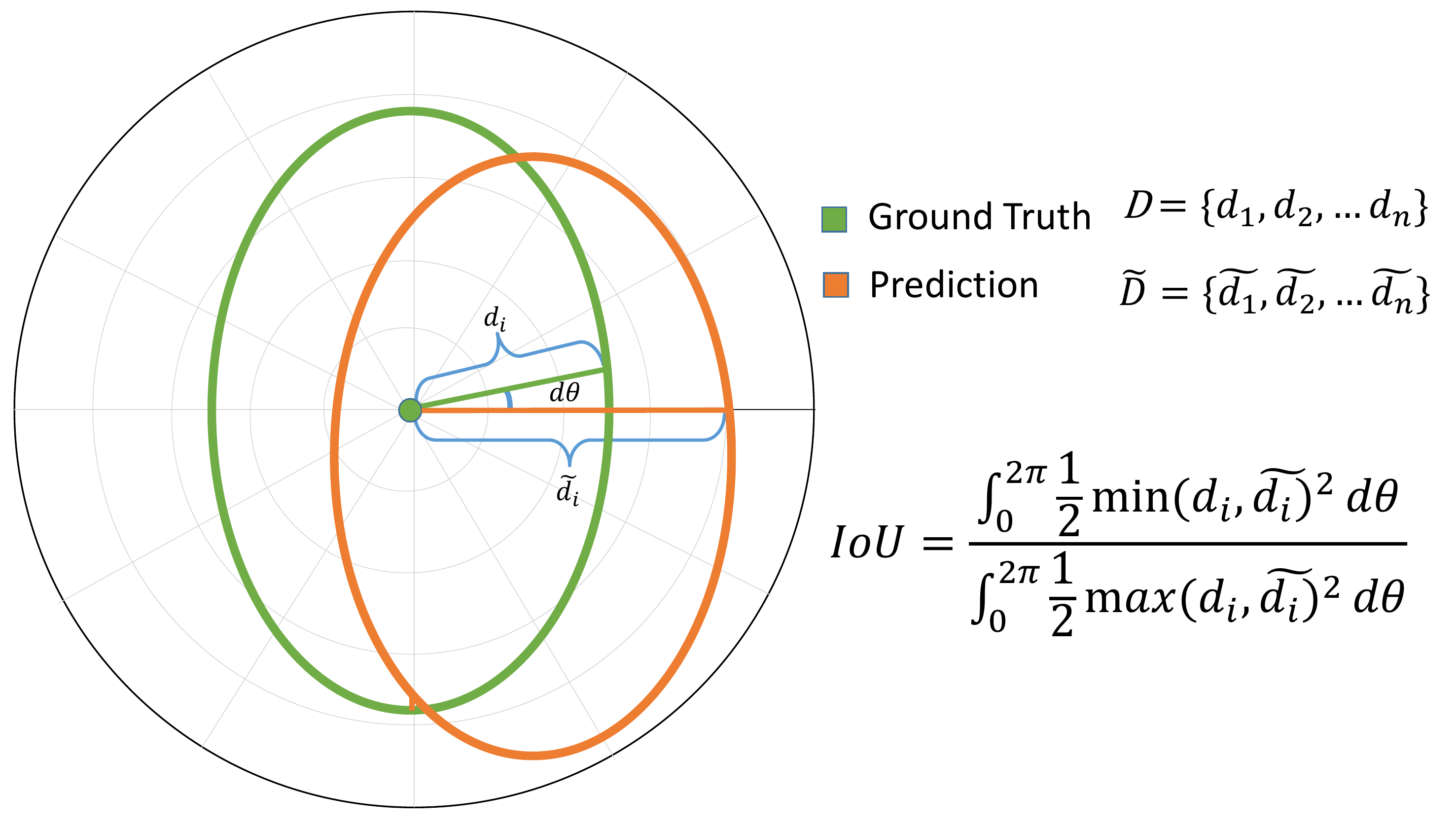}
\caption{\textbf{Mask IoU in Polar Representation}. Mask IoU (interaction area over union area) in the polar coordinate can be calculated by integrating the differential IoU area in terms of differential angles. Polar IoU loss optimizes the mask regression as a whole, instead of optimizing each ray separately like L1 loss, leading to better performance.}
\label{fig_iouloss}
\vspace{-5mm}
\end{figure}

Intuitively, optimization of the polar IoU loss in Eqn.\ref{IoULoss} encourages lengths of the predicted rays to be the same as the target rays.
It is derived from the polar mask IoU introduced below.
Here we connect the polar IoU to the polar mask IoU that has a continuous formation, showing that Eqn.\ref{IoULoss} is actually maximizing the mask IoU in the polar space. The polar mask IoU is the ratio between the predicted mask and the ground-truth mask represented by the polar coordinates.
As shown in Figure~\ref{fig_iouloss}, mask IoU can be calculated by using polar integration
\begin{equation}
  { \rm IoU  } = \frac{\int_{0}^{2\pi} \frac{1}{2} \min(d,d^*)^2 d\theta}
        {\int_{0}^{2\pi}\frac{1}{2} \max(d,d^*)^2 d\theta},
\end{equation}
where $d$ and $d^*$ are the target and the predicted lengths of the rays respectively and $\theta$ is the angle between rays.
We can transform the above continuous form into a discrete form,
\begin{equation}
 {  \rm IoU } = \lim_{N \to \infty}\frac
  {\sum_{i=1}^{N}\frac{1}{2} d_{\min}^2 \Delta \theta_i}
  {\sum_{i=1}^{N}\frac{1}{2} d_{\max}^2 \Delta \theta_i}.
\end{equation}
In fact, when $N$ approaches infinity, this discrete form equals the continuous form. 
We assume that the rays are uniformly emitted so that we have $\Delta \theta = \frac{2\pi}{N}$, which can further simplify the expression. 
In practice, we empirically observe that the power of two has a negligible impact on the performance~($\pm0.1$ mAP on COCO).
Thus, we ignore the power of two and apply the definition in Eqn.~\ref{PolarIoU} to calculate the Polar IoU to approximate the mask IoU.

\subsection{Network Architecture and Model Training \label{sec:Architecture}}

The proposed system PolarMask++ is an effective and unified framework, which consists of a backbone network, a  modified feature pyramid network, and the task-specific heads.
To enable fair comparisons, the setup of the backbone follows FCOS~\cite{FCOS}, which is a representative method for one-stage object detection. 
Although there are many candidates for the backbone networks, we align the setting with FCOS to show the simplicity and effectiveness of our instance modeling method.
As shown in the middle of Figure~\ref{fig:pipeline}, the  heads in PolarMask++ contain three branches, including a classification branch, a polar centerness branch, and a mask regression branch, which predict the class label, the polar centerness score and the length of each polar ray of each pixel respectively, where
$k$ and $n$ indicate the number of categories and the number of rays.

\subsubsection{Refined Feature Pyramid \label{rfp}}
Feature Pyramid Network~(FPN)~\cite{fpn} is commonly used in most object detection and instance segmentation methods such as RetinaNet and Mask R-CNN. FPN has achieved great success because high-level features in backbones have more semantic meanings while shallow low-level features have more content information. 
In the literature, the Balanced Feature Pyramid~(BFP) is proposed in~\cite{pang2019libra} balance the feature in each resolution in object detection.
In this paper, we propose a Refined Feature Pyramid (RFP) that strengthens the feature pyramid representation in instance segmentation.

\begin{figure}[!t]
\begin{center}
\includegraphics[width=0.49\textwidth]{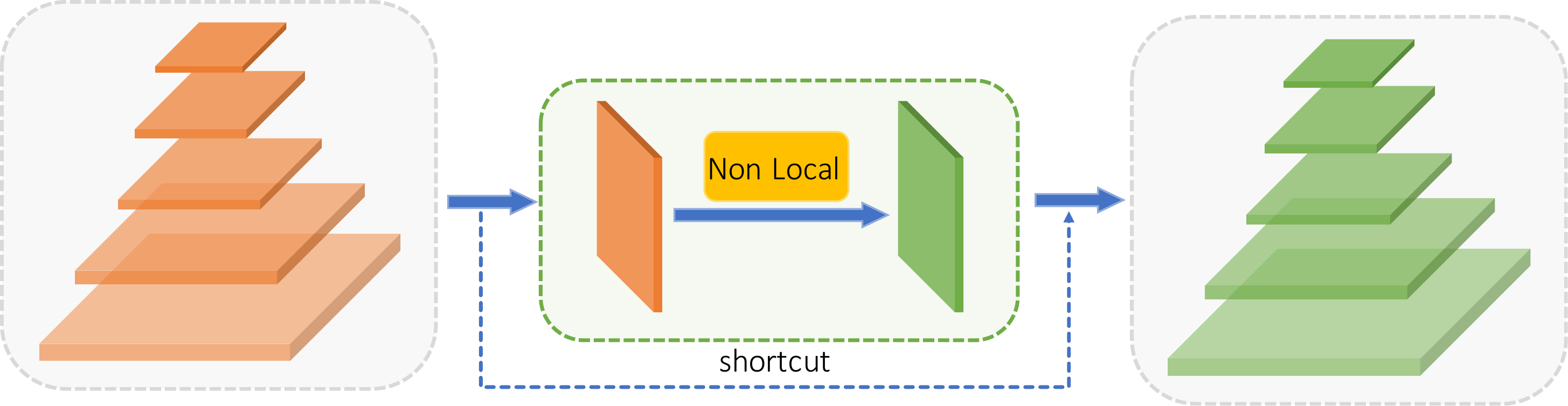}
\caption{\textbf{Refined Feature Pyramid}. The feature maps from different stages are integrated into the same size and added, then a non-local block is used for context modeling. At last, the refined feature is integrated to original sizes and a shortcut operate is adopted to get the final feature representation in multiple scales.}
\label{fig:rfp}
\end{center}
\vspace{-5mm}
\end{figure}

As shown in Figure~\ref{fig:rfp}, given the feature maps from $P3, P4, P5, P6, P7$ with resolution range from $1/8$ to $1/128$, 
we first re-scale the feature maps in different levels to the resolution of $1/8$,
and then fuse them by adding all these feature maps. 
Next, we use non-local~\cite{nonlocal} operation on the fused feature maps to calculate the relationship between long-range and short-range pixels and ``refine'' the representation of each pixel by using its contextual information.
After we get the refined feature maps, we re-scale these feature maps using a similar but reverse procedure. 
Finally, to generate the final feature representation, the origin feature maps from multiple scales are added with the refined feature maps by using a shortcut connection. 

The Refined Feature Pyramid not only makes high-resolution features and low-resolution features fused more effectively but also builds the relation of pixels on the feature map. 
In experiments, we find that it benefits object segmentation, especially for small instances.

\subsubsection{Label Generation \label{labelgen}}
Here we explain the procedure of  label generation for ray regression in Eqn.~\ref{IoULoss},
as shown in Algorithm~\ref{alg:distance}. 
Firstly, we  obtain the contours of one instance by applying methods such as $\tt cv2.findContours$ in OpenCV. 
Secondly, we traverse every point on the contour to calculate the distance and the angle from this point to the center of object instance. 
Finally, we could achieve the distance given  the  corresponding angle 
(\eg when number of rays is 36, we obtain ray length of every $\Delta\theta=10^{\circ}$). 
When the target angle is missing, we adopt the nearest angle as the supervision. 
For instance, if the ray of $10^{\circ}$ is missing but that corresponding to $9^{\circ}$ exists, 
we can use $9^{\circ}$ as the regression target.

\subsubsection{Model Training \label{optimize}}
The proposed model optimizes multiple tasks jointly including label prediction, polar centerness prediction and polar ray regression. We therefor define a multi-task loss function,
\begin{equation}
L = L_{cls} + \alpha_1 L_{reg} + \alpha_2 L_{ct},
\end{equation}
where $L_{cls}$ is the classification loss that is formulated as the focal loss function.
$L_{reg}$ is the mask regression loss, which is defined in section~\ref{iouloss}. 
And $L_{ct}$ is the loss for soft polar centerness, which is formulated as a binary cross entropy loss.
We set the trade-off parameters $\alpha_1$ and $\alpha_2$ to `1' in all experiments.

\begin{table*}[t]
    \centering
    \scalebox{1.0}{
    \def\x{{$\footnotesize \times$}}
\begin{tabular}{l|c|l|cc|ccc|ccc|c}
	method & venue & backbone & epochs & aug & AP & AP$_{50}$ & AP$_{75}$ & AP$_{S}$ & AP$_{M}$ & AP$_{L}$ & FPS\\
	\hline
    \emph{two-stage} &&&&&&&&&&\\
    MNC~\cite{mnc} &CVPR'16 & ResNet-101 & 12 & $\circ$ & 24.6 & 44.3 & 24.8 & 4.7 & 25.9 & 43.6 & $<$1\\
	FCIS~\cite{fcis} & CVPR'17 & ResNet-101 & 12 & $\circ$ & 29.2 & 49.5 & - & 7.1 & 31.3 & 50.0 & $<$1\\
	Mask R-CNN~\cite{maskrcnn} & ICCV'17 & ResNeXt-101 & 12 &$\circ$ & 37.1 & 60.0 & 39.4 & 16.9 & 39.9 & 53.5 & 8\\
	\hline
	\emph{one-stage} &&&&&&&&&&\\
	ExtremeNet~\cite{extremenet} & CVPR'19 & Hourglass-104 & 100 &\checkmark &  18.9  & 44.5 & 13.7 & 10.4 & 20.4 & 28.3 &3 \\
	TensorMask~\cite{tensormask} & ICCV'19 & ResNet-101 & 72 & \checkmark & 37.1 & 59.3 & 39.4 & 17.1 & 39.1 & 51.6 &3 \\
	YOLACT~\cite{yolact} & ICCV'19 & ResNet-101 & 48 &\checkmark &  31.2  & 50.6 & 32.8 & 12.1 & 33.3 & 47.1 & 23 \\
	PolarMask~\cite{polarmask} & CVPR'20 & ResNet-101 & 24 & \checkmark &32.1  &53.7  &33.1  &14.7  & 33.8 &45.3 &15 \\
	PolarMask~\cite{polarmask} & CVPR'20 & ResNeXt-101-DCN & 24 & \checkmark &36.2  &59.4 &37.7 &17.8  &37.7  &51.5 &7\\
	
	\textbf{PolarMask++(600)} & - & ResNet-101 & 24 & \checkmark &32.3 &55.0 &33.3 &13.2 &34.1 &47.8 &\textbf{24}\\
	\textbf{PolarMask++} & - & ResNet-101 & 24 & \checkmark      &33.8 &57.5 &34.6 &16.6 &35.8 &46.2 &14\\
	\textbf{PolarMask++} & - & ResNeXt-101-DCN & 24 & \checkmark &37.2  &62.3 &38.5 &19.6  &39.2  &51.4 &6\\
	\textbf{PolarMask++$^{*}$} & - & ResNeXt-101-DCN & 24 & \checkmark &38.7  &64.1 &40.0 &22.2  &40.2  &52.0 &4\\
\end{tabular}
}
    \vspace{2mm}
	\caption{\textbf{Instance segmentation} mask AP on the COCO $\tt test$-$\tt dev$. The standard training strategy~\cite{Detectron} is training by 12 epochs; and `aug' means data augmentation, including multi-scale and random crop. $\checkmark$ is training with `aug', $\circ$ is without `aug'.
	`PolarMask++(600)' means we scale the short side of test image to 600 for faster speed. Our PolarMask++(600) achieves better speed and performance than real-time instance segmentation method YOLACT.}
	$*$ indicates we enlarge the input input image to $1920\times1280$. It can further boost up the performance, especially helpful for small object segmentation. And the final best performance is higher than Mask R-CNN, showing our method is competitive on general instance segmentation task.
	\label{table:2}
\end{table*}

\begin{table*}[t]
    \centering
    \scalebox{1.0}{
    \def\x{{$\footnotesize \times$}}
\begin{tabular}{l|c|c|c|c|ccc|c}
			method & venue &representation & backbone & external data  & Precision & Recall & F-measure & FPS\\
			\hline
			CTPN~\cite{tian2016detecting}  & ECCV'16 &cartesian &VGG16 &-  & 74.2 & 51.5 & 60.8 & 7.1 \\
			EAST~\cite{east}               & CVPR'17 &cartesian &VGG16 & -  & 83.5 & 73.4 & 78.2 & 13.2 \\
			RRPN~\cite{rrpn}               & TMM'18  &cartesian &VGG16 &-  & 82.0 & 73.0 & 77.0 & - \\
			DeepReg~\cite{deepreg}         & ICCV'17 &cartesian &VGG16 & -  & 82.0 & 80.0 & 81.0 & -\\
			PixelLink~\cite{PixelLink}     & AAAI'18 &pixel &VGG16 & -  & 82.9 & 81.7 & 82.3 &7.3\\
			PAN~\cite{pan}                 & ICCV'19 &pixel &ResNet-18 & -  & 82.9 & 77.8 & 80.3 & \textbf{26.1} \\
			\hline
			\textbf{PolarMask++}           & -       &polar &ResNet-50 & -  & 86.2 & 80.0 & \textbf{83.4} & 10\\
			\hline
			\hline
			SegLink~\cite{shi2017detecting}& CVPR'17 &cartesian &VGG16 &\checkmark  & 73.1 & 76.8 & 75.0 & - \\
			SSTD~\cite{he2017single}       & ICCV'17 &cartesian &VGG16 &\checkmark  & 80.2 & 73.8 & 76.9 & 7.7 \\
			WordSup~\cite{hu2017wordsup}   &CVPR'17  &pixel &VGG16 &\checkmark  & 79.3 & 77.0 & 78.1 & -  \\
			Lyu et al.~\cite{lyu2018multi} &CVPR'18  &pixel &VGG16 & \checkmark  & 94.1 & 70.7 & 80.7 & 3.6 \\
			RRD~\cite{rrd}                 &CVPR'18  &cartesian &VGG16 &\checkmark  & 85.6 &79.0 &82.2 &6.5\\
			MCN~\cite{mcn}                 &CVPR'18  &pixel &VGG16 &\checkmark  & 72.0 &80.0 &76.0 &-\\
			TextSnake~\cite{textsnake}     &ECCV'18  &pixel &VGG16 & \checkmark  & 84.9 & 80.4 & 82.6 & 1.1 \\
			PSENet~\cite{psenet}           &CVPR'19  &pixel &ResNet-50 & \checkmark  & 86.1 & 83.7 & 84.9 & 3.8 \\
			MSR~\cite{msr}                 &IJCAI'19 &pixel &ResNet-50 & \checkmark  & 86.6 & 78.4 & 82.3 & 4.3 \\
			TextField~\cite{textfield}     &TIP'19   &pixel &VGG16  & \checkmark  & 84.3 & 80.5 & 82.4 & 5.2 \\
			PAN~\cite{pan}                 & ICCV'19 &pixel &ResNet-18 & \checkmark  & 84.0 & 81.9 & 82.9 & \textbf{26.1} \\
			TextDragon~\cite{textdragon}   & ICCV'19 &cartesian &VGG16 & \checkmark  & 84.8 & 81.8 & 83.1 & - \\
			\hline
			\textbf{PolarMask++}           & -       &polar &ResNet-50 & \checkmark  & 87.3 & 83.5 & \textbf{85.4} & 10\\

	\end{tabular}}
    \vspace{2mm}
	\caption{The single-scale results on rotate object detection dataset ICDAR2015.  Following the COCO setup and without using any tricks, PolarMask++ achieves highest F-measure with or without external data, showing polar representation is suitable for scene text detection and rotate object detection.}
	\label{table:ic15}
\end{table*}

\begin{table*}[t]
    \centering
    \scalebox{0.95}{
    \def\x{{$\footnotesize \times$}}
\begin{tabular}{l|c|c|cccccccccc}
	method & venue                       & representation & AP$_{50}$ & AP$_{55}$ & AP$_{60}$ & AP$_{60}$ & AP$_{70}$ & AP$_{75}$ & AP$_{80}$ & AP$_{85}$ & AP$_{90}$ & mAP\\
	\hline
	DCAN~\cite{dcan}              &CVPR'16          & pixel & - & - & - & - & - & - & - & - & - & 51.1 \\
    U-Net~\cite{unet}             &MICCAI'16        & pixel &80.6&77.5&74.3&70.1&65.4&57.7&49.1&37.3&22.5&59.8\\
	Mask R-CNN~\cite{maskrcnn}    & ICCV'17         & pixel &83.2&80.5&77.2&72.9&68.3&59.7&48.9&35.2&18.9&60.5\\
	StarDist~\cite{celldet}       & MICCAI'18       & polar &86.4&83.6&80.4&75.4&68.5&58.6&44.9&28.6&11.9&59.8 \\
	Keypoint Graph~\cite{kpgraph} & MICCAI'19       & pixel &71.5&-&-&59.3&-&-&-&-&-&- \\
	Nuclei R-CNN~\cite{nrcnn}     & ICICSP'19       & pixel & - & - & - & - & - & - & - & - & - & 69.6 \\
	Li \textit{et al}~\cite{yi2019object} & AAAI'20 & pixel & 84.8 & - & - & - & - & 65.1 & - & - & - & 61.1\\
	PatchPerPix~\cite{patchperpix}        & ECCV'20 & pixel &87.1&-&82.4&-&75.2&-&60.9&-&\textbf{35.2}&68.1\\
	\hline
	\textbf{PolarMask++}                  & - & polar &\textbf{92.7}&\textbf{91.2}&\textbf{89.6}&\textbf{87.3}&\textbf{84.0}&\textbf{78.3}&\textbf{68.0}&\textbf{50.7}&25.8&\textbf{74.2}\\
	
\end{tabular}
}
    \vspace{2mm}
	\caption{The single-scale results on cell segmentation dataset DSB2018. 
	Following the COCO setup and without using any tricks, PolarMask++ achieves highest mAP, making 4.6 mAP higher than previous best method. It shows polar representation has large advantages on cell instance segmentation task.}
	Note that the result of U-Net and Mask R-CNN are Obtained from StarDist\cite{celldet}.
	\label{table:dsb}
\end{table*}

\section{Experiments}
To validate the effectiveness of the proposed approaches, we conduct extensive experiments and compare with the recent state-of-the-art methods on three challenging public benchmarks, including a general instance segmentation dataset MSCOCO~\cite{coco},  a rotated object detection dataset for text detection ICDAR2015~\cite{2015icdar} and a cell instance segmentation dataset DSB2018~\cite{dsb}.
All the experiments are implemented in mmDetection~\cite{mmdetection} using PyTorch. All networks are trained with 8 NVIDIA Tesla V100 GPUs with 32GB memories.

\begin{algorithm}[t]
		\footnotesize 
		\caption{Distance Label Generation~(36 rays)}
		\begin{algorithmic}[1]
			\Require Contour: $Contour$, Center Sample: $center$, 
			\Function {Distance Calculate}{$Contour$, $center$}
			\State Initialize distance set D, angle set A
			\For{each $point \in Contour$}
			    \State Calculate distance and angle from $point$ to $center$
			    \State Append distance to D, angle to A
			\EndFor
			\State Get distance set D, angle set A  
			\\
			\State Initialize distance label L$_D$
            \For{angle $\theta$ $\in$ [0,10,20,\dots,360]}
                \If {Find angle $\theta$ in A}
                    \If {angle has multiple distances $d$}  
                        \State Find the maximum $d$
                    \Else
			            \State Find corresponding distance $d$
			        \EndIf
    			\Else {~~$\theta$ not in A}
    			    \If {Find angle $\theta_{near}$ nearby $\theta$ in A}
    			        \State Find corresponding $d$    \ \ \ \ \  // Nearest Interpolation.
    			    \Else
    			        \State $d$ = $10^{-6}$        \ \ \ \ \ \ \ \ \ \ \ \ \ // Target a minimum number as label.
    			    \EndIf
    			\EndIf
    			\State Append $d$ to L$_D$
    		\EndFor
    		\State \Return{L$_D$}
			\EndFunction
			
		\end{algorithmic}
		\label{alg:distance}
	\end{algorithm}
	
\subsection{General Instance Segmentation on COCO}

\textbf{Experiment Settings.} We first examine the performance of the proposed PolarMask++ on the COCO benchmark~\cite{coco}, which is a widely used dataset in general object detection and instance segmentation.
In COCO, Average Precision~(AP) is used to measure the performance.
By following common setups~\cite{maskrcnn,tensormask}, we train the models by using the union of 80K training images and a 35K subset of validation images ($\tt trainval35k$), and report results on the remaining 5K validation images ($\tt minival$). We also compare our results on $\tt test$-$\tt dev$ with the recent state-of-the-art methods, including both the one-stage and two-stage models.

Similar to~\cite{maskrcnn,tensormask}, we employ ResNet101 and ResNeXt101 as the backbone networks of PolarMask++. 
In the training phase, we adopt 2$\times$ training schedule~(\ie 24 epochs).
All the models are trained with 4 samples per GPU and are optimized by using stochastic gradient decent (SGD) with an initial learning rate of 0.02.
All the input images are resized to 768$\times$1280 to input the network.
For data augmentation, we randomly scale the shorter side of images in the range from 640 to 768 during the training. 
During inference, we keep the input size of 768$\times$1280 for single scale testing unless otherwise stated. 

\noindent \textbf{Result Comparisons.} Table \ref{table:2} reports the performance of PolarMask++ against its counterparts including the recent one-stage and two-stage models. Without bells and whistles, PolarMask++ is able to achieve competitive performance with more complex one-stage methods. For example, using a simpler pipeline and half training epochs, PolarMask++ outperforms YOLACT by 2.6 mAP. Moreover, PolarMask++ with deformable convolutional layers~\cite{dcn} can achieve 37.2 mAP, which is comparable with state-of-the-art methods.
We further enlarge the input image from $768\times1280$ to $1280\times1920$ and the performance consistently improved, especially the AP for small objects. This is because there are many small objects~(41\% objects) in COCO, so a high-resolution input image is helpful for small objects. In this case, the best PolarMask++ can achieve 38.7 mAP, outperforming the prior state-of-the-art two-stage method Mask R-CNN by 1.6\%. 
We compare the runtime (\ie frame per second, FPS) between TensorMask and PolarMask++ with the same image size~(short length 800) and device~(one V100 GPU). PolarMask++ runs at 14 FPS with the ResNet-101 backbone, which is more than 4 times faster than TensorMask~(3 FPS).

\begin{figure*}[h!]
\begin{center}
\includegraphics[width=0.85\textwidth]{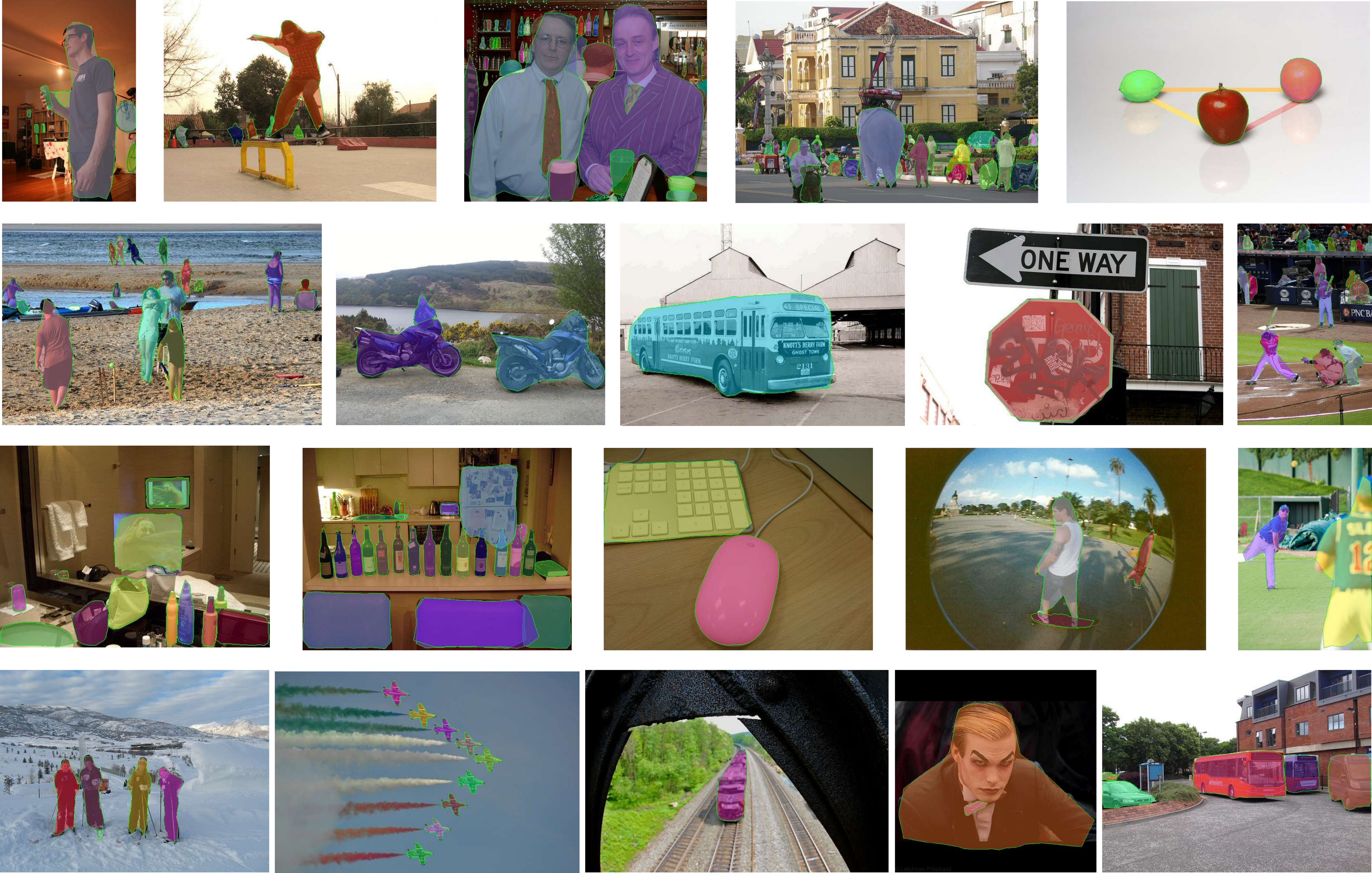}
\vspace{2mm}
\caption{Example results of PolarMask++ on COCO $\tt test$-$\tt dev$ using ResNeXt-101-DCN as backbone, which achieves 37.2\%  mask AP (see Table~\ref{table:2}). Zoom in for best view.}
\label{fig:result}
\end{center}
\vspace{2mm}
\end{figure*}

The outputs of PolarMask++ are visualized in Figure~\ref{fig:result}, whereby we have two observations.
(1) For objects with regular shapes such as bus and apple, PolarMask++ predicts accurate contours than other methods.
(2) For objects with complex and non-regular shapes such as person, PolarMask++ predicts relatively rough contours, while the performance is not satisfactory enough. These objects make PolarMask++ have relatively low performance under high-IoU restriction.
For instance, in Table~\ref{table:2},  PolarMask++ achieves 64.1\% in AP$_{50}$, improving 4.1\% compared to Mask R-CNN. However, in AP$_{75}$, the improvement over Mask R-CNN is just 0.6\%.
We would like to point out that it is the main challenge of polar representation on instance segmentation, and we will put more effort to solve it.

\subsection{Rotated Text Detection}
\textbf{Experimental Settings.} We also evaluate the performance
of PolarMask++ on the rotated detection task of ICDAR2015 \cite{2015icdar}. This dataset deals with scene text detection, a typical rotated detection task.
It contains $1000$ training samples and $500$ test images. All training images are annotated with word-level quadrangles as well as corresponding transcriptions. In ICDAR2015, F-measure, which is the Harmonic mean of precision and recall, is used as the metric.

Following the common setting~\cite{spcnet,psenet,msr}, we use ResNet-50 as the backbone network for fair comparisons. Similar to prior arts, the COCO pre-trained model is used to initialize the parameters of the network. We report two kinds of results, including (1) do not use external text detection dataset to pre-train the network; (2) follow \cite{spcnet,psenet,pan} to use external text dataset ICDAR2017 to pre-train the network.

In training, the batch size is 4 images per GPU and a multi-scale strategy is adopted to keep the short edge from 768 to 1024.
All models are trained by using SGD optimizer for 48 epochs with the initial learning
rate as 0.02.
The learning rate is divided by 10 at the 32-th and the 44-th epoch.
In the test, the short-side of the image is 1024 and we perform the single-scale testing process. We do a grid search then set the NMS threshold 0.3 and the score 0.25 to balance the Precision and Recall.

\noindent\textbf{Result Comparisons.}   We compare our method with the recent state-of-the-art methods as shown in Table~\ref{table:ic15}. 
With only single-scale testing, our method achieves 83.4\% F-measure without external text dataset for pre-training, outperforming prior best-performing method by 1\%. When an external dataset is used to pre-train the backbone, PolarMask++ achieves 85.4\%, establishing a new state-of-the-art performance. 
With the only post-process NMS, PolarMask++ can run at 10 FPS, leading to large advantages in speed compared to most of the methods such as PSENet and TextSnake, which typically need more complex pipelines and post-processing steps. 
The above competitive results on ICDAR2015 verify that PolarMask++ can not only perform instance segmentation but also transfer to rotated object detection, which demonstrates the advantages of polar representation.

In addition, we demonstrate some detection examples in appendix,  showing that PolarMask++ can accurately detect arbitrarily oriented text instances, which proves that the polar representation can handle not only instance segmentation but also rotated object detection.

\subsection{Cell Segmentation}

\textbf{Experimental Settings.} To further evaluate the robustness of PolarMask++, we conduct experiments on the DSB2018 dataset, 
which is a cell dataset manually annotated real microscopy images of cell nuclei from the 2018 Data Science Bowl. It contains 671 images. The cells in this dataset have different types,  magnification, and imaging modalities because they were captured under different conditions.
This dataset does not provide training and testing partitions.  By following the prior work~\cite{celldet}, we randomly use 90\% of the images for training and 10\% for testing. 
In DSB2018, the evaluation metric is the same as COCO, we use AP to evaluate the performance.

We use ResNet-50 as the backbone network. The COCO pre-trained model is used to initialize the parameters of the network. In training, 
the batch size is 4 images per GPU and a multi-scale strategy is adopted to keep the short edge of the image from 768 to 1024.
All models are trained for 12 epochs by using SGD optimizer with an initial learning rate of 0.02.
The learning rate is divided by 10 at the 8-th and 11-th epoch.
In the test, the short-side of the image is 1024 and single-scale testing is performed. 
In DSB2018, as the number of cells may be large, the max objects per image are 400 during inference. By following the previous work, we report AP from AP$_{50}$ to AP$_{90}$ and the mAP as the metric.

\noindent\textbf{Result Comparisons.}
The results on DSB2018 are shown in Table~\ref{table:dsb}. 
Without bells and whistles, PolarMask++ clearly outperforms its counterparts with large margins.
For example, PolarMask++ surpasses Nuclei R-CNN, which is the recent state-of-the-art method, by 4.6\% mAP. It also has large advantages over Mask R-CNN, which is a state-of-the-art instance segmentation method. We see that the polar representation is able to perform accurately cell segmentation. 

Some results are visualized in appendix. We have two observations.
(1) Polar representation has natural advantages to represent cells because cell tends to have regular shapes. 
(2) PolarMask++ is robust to detect cells with different backgrounds and scales.
We believe PolarMask++ can be immediately applied in the industry with a simple scenario such as cell segmentation and corn detection.

\subsection{Summarizing the Experimental Results}
The above experimental analyses of general instance segmentation rotated object detection and cell instance segmentation in various benchmarks can be concluded as follows.
(1) Polar representation is a flexible and general representation that could model instance, rotated object as well as cell. It has great advantages compared to its counterparts.
(2) The proposed method has a huge superiority on rotated object detection and cell instance segmentation, and the runtime speed is fast.
(3) Compared to the prior PolarMask~\cite{polarmask}, the proposed method achieves much better performance, which strongly proves the effectiveness of soft polar centerness and refined feature pyramid.

\subsection{Advantages of PolarMask++ \textit{vs.} Mask R-CNN}
\begin{figure}[t]
\begin{center}
\includegraphics[width=0.47\textwidth]{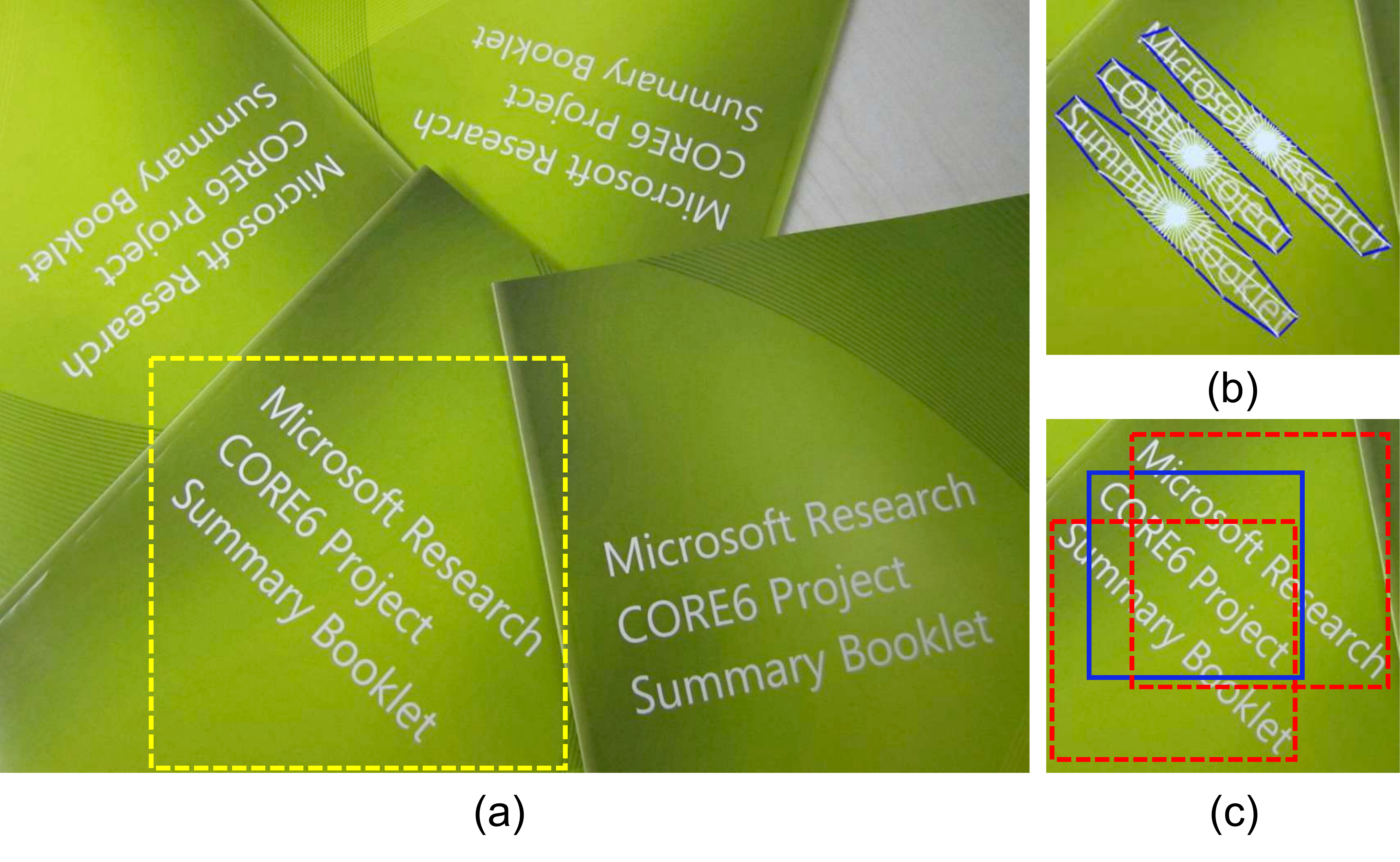}
\caption{The advantage of PolarMask++ on heavy crowd and rotate scene compared with Mask R-CNN. (a) shows a representative original image with large rotations from a text detection dataset MSRA-TD500~\cite{msra}. (b) Detection result of PolarMask++. (c) Detection result of Mask R-CNN. The blue solid box and red dot boxes in (c) are the remained and suppressed results by NMS. 
It shows that polar representation has more advantages than bounding boxes in such situation, because bounding boxes are not tied and the highly-overlapped boxes are easily suppressed by NMS, leading to mis-detected instance. However, our PolarMask system directly predicts center point and ray lengths of object in the polar coordinate without relying on bounding box, thus being suitable in challenging situations such as heavy rotation and crowded objects.}
\label{fig:polar_advantage}
\end{center}
\vspace{-5mm}
\end{figure}

The PolarMask++ has obviois advantages compared with Mask R-CNN in a crowded scene. Mask R-CNN is a classic bounding-box-detection based method. As shown in Figure~\ref{fig:polar_advantage}, when detecting dense objects, Mask R-CNN needs to detect object's bounding box first. However, the bounding boxes of crowd objects are highly overlapped, which will be suppressed by non-maximum suppression (NMS). 
Moreover, although Mask R-CNN successfully detects the bounding box, there tend to be more than one object in one bounding box, which is also challenging for mask segmentation.
However, the advantage of PolarMask++ is that it directly regress the boundary of object from the object center, and does not rely on box detection. In this way, as long as the centers of dense objects do not overlap, PolarMask++ is able to detect these objects and not miss one object easily.

\begin{table*}[t]	
	\centering
\begin{subtable}[t]{3.2in}
	\centering
	\setlength{\tabcolsep}{1.5mm}
	\small 
	\begin{tabular}{c|ccc|ccc}
	rays & AP & AP$_{50}$ & AP$_{75}$ & AP$_{S}$ & AP$_{M}$ & AP$_{L}$ \\
	\hline
	18 & 26.2 & 48.7 & 25.4 &11.8 &28.2 &38.0\\
	24 & 27.3 & 49.5 & 26.9 &12.4 &29.5 &40.1\\
	\textbf{36} & \textbf{27.7} &49.6  &  27.4 &  12.6 & 30.2 & 39.7\\
	72 & 27.6 &  49.7 & 27.2 & 12.9 &30.0 & 39.7\\
	\end{tabular}
	\caption{\textbf{Number of Rays}: More rays bring a large gain, while too many rays saturate since it already depicts the mask ground-truth well.}
	\label{table:vectors}
	\vspace{5mm}
\end{subtable}
\quad
\begin{subtable}[t]{3.2in}
    \centering
	\setlength{\tabcolsep}{1.0mm}
	\small 
	\begin{tabular}{c|c|ccc|ccc}
	loss & $\alpha$  & AP & AP$_{50}$ & AP$_{75}$ & AP$_{S}$ & AP$_{M}$ & AP$_{L}$\\
	\hline
	\multirow{3}*{Smooth-\L1} &
	0.05 & 24.7 & 47.1 & 23.7 &11.3 &26.7 &36.8\\
	& 0.30 & 25.1 & 46.4 & 24.5 &10.6 &27.3 &37.3\\
	& 1.00 & 20.2 & 37.9 & 19.6 &8.6 &20.6 &31.1\\
	\hline
	\textbf{Polar IoU} & 1.00  & \textbf{27.7} & 49.6 & 27.4 &12.6 &30.2 &39.7 \\
	\end{tabular}
	\caption{\textbf{Polar IoU Loss \emph{vs}. Smooth-L1 Loss}: Polar IoU Loss outperforms Smooth-\L1 loss, even the best variants of balancing regression loss and classification loss.}
	\label{table:loss}
	\vspace{5mm}
\end{subtable}
\quad
\begin{subtable}[t]{3.2in}
	\centering
	\setlength{\tabcolsep}{1.2mm}
	\small 
	\begin{tabular}{c|ccc|ccc}
	centerness & AP & AP$_{50}$ & AP$_{75}$ & AP$_{S}$ & AP$_{M}$ & AP$_{L}$\\
	\hline
	Original & 27.7 & 49.6 & 27.4 & 12.6&30.2& 39.7\\
	\textbf{Polar} & \textbf{29.1} & 49.5 & 29.7 & 12.6 & 31.8 & 42.3\\
	\end{tabular}
	\caption{\textbf{Polar Centerness \emph{vs}.  Centerness}: Polar Centerness bring a large gain, especially high IoU AP$_{75}$ and large instance AP$_L$.}
	\label{table:centerness}
	\vspace{5mm}
\end{subtable}
\quad
\begin{subtable}[t]{3.2in}
	\centering
	\setlength{\tabcolsep}{1.2mm}
	\small 
	\begin{tabular}{c|ccc|ccc}
	box branch & AP & AP$_{50}$ & AP$_{75}$ & AP$_{S}$ & AP$_{M}$ & AP$_{L}$\\
	\hline
	w & 27.7 & 49.6 & 27.4 & 12.6&30.2&39.7\\
	w/o & 27.5 & 49.8 & 27.0 &13.0 &30.0 &40.0\\
	\end{tabular}
	\caption{\textbf{Box Branch}: Box branch makes no difference to performance of mask prediction.}
	\label{table:box}
	\vspace{5mm}
\end{subtable}
\quad
\begin{subtable}[t]{3.2in}
	\centering
	\setlength{\tabcolsep}{1.5mm}
	\small 
	\begin{tabular}{l|ccc|ccc}
	centerness & AP & AP$_{50}$ & AP$_{75}$ & AP$_{S}$ & AP$_{M}$ & AP$_{L}$\\
	\hline
	Polar        & 29.1 & 49.5 & 29.7  & 12.6 & 31.8 & 42.3 \\
	Soft~($F_1$) & 29.7 & 51.6 & 30.4 & 13.3 & 32.0 & 42.7 \\
	Soft~($F_2$) & 29.7 & 51.6 & 30.0 & 13.9 & 31.7 & 42.9 \\
	Soft~($F3$) & 29.8 & 52.0 & 30.2 & 13.9 & 31.9 & 43.0 \\
	\end{tabular}
	\caption{\textbf{Soft Mechanism}: Compare with conference version~\cite{polarmask}, soft polar centerness can improve the performance without computation overhead.}
	\label{table:soft}
\end{subtable}
\quad
\begin{subtable}[t]{3.2in}
	\centering
	\setlength{\tabcolsep}{1.1mm}
	\small 
	\begin{tabular}{l|ccc|ccc|cc}
	neck & AP & AP$_{50}$ & AP$_{75}$ & AP$_{S}$ & AP$_{M}$ & AP$_{L}$ & Flops & Params\\
	\hline
	FPN & 29.8 & 52.0 & 30.2 & 13.9 & 31.9 &43.0 &252.5 &34.4\\
	PAN & 29.9 & 51.8 & 30.6 & 13.5 & 32.1 & 43.0 &258.5 &36.8\\
	RFP & 30.2 & 52.6 & 30.8  & 14.4 & 32.5 & 43.1 &253.6 &34.7\\
	\end{tabular}
	\caption{\textbf{Strategy of Feature Pyramid}: Compare with conference version~\cite{polarmask} and PAN~\cite{panet}, our proposed RPF can consistently boost up the performance, especially for small objects.
	}
	\label{table:rfp}
\end{subtable}
\quad
\begin{subtable}[t]{3.2in}
	\centering
	\setlength{\tabcolsep}{1.0mm}
	\small 
	\begin{tabular}{l|ccc|ccc}
	backbone & AP & AP$_{50}$ & AP$_{75}$ & AP$_{S}$ & AP$_{M}$ & AP$_{L}$\\
	\hline
	ResNet-50 & 30.2 & 52.6 & 30.8  & 14.4 & 32.5 & 43.1 \\
	ResNet-101 & 31.6 & 54.5 & 32.2 & 15.7 & 34.2 & 44.8 \\
	ResNeXt-101 & 33.9 & 57.7 & 34.9 & 17.3 & 37.0 & 47.7 \\
	\end{tabular}
	\caption{\textbf{Backbone Architecture}: All models are based on FPN. Better backbones bring expected gains: deeper networks do better, and ResNeXt improves on ResNet.}
	\label{table:backbone}
\end{subtable}
\quad
\begin{subtable}[t]{3.2in}
	\centering
	\setlength{\tabcolsep}{1.0mm}
	\small 
	\begin{tabular}{l|ccc|ccc|c}
	scale & AP & AP$_{50}$ & AP$_{75}$ & AP$_{S}$ & AP$_{M}$ & AP$_{L}$ &  FPS\\
	\hline
	400 & 24.0 & 40.8 & 24.4  & 5.7 & 25.2 & 43.0 & 48.8\\
	600 & 28.5 & 48.5 & 29.2 & 10.7 & 31.0 & 44.2 & 33.2 \\
	800 & 30.2 & 52.6 & 30.8  & 14.4 & 32.5 & 43.1 & 20.5\\
	\end{tabular}
	\caption{\textbf{Accuracy/speed trade-off on ResNet-50}: PolarMask performance with different image scales. The FPS is reported on one V100 GPU.
	}
	\label{table:speed}
\end{subtable}

	\caption{Ablation experiments for the proposed PolarMask++ on MSCOCO dataset. All models are trained on $\tt trainval35k$ and tested on $\tt  minival$, using ResNet50-FPN backbone with 1$\times$ training schedule unless otherwise noted.}
	\label{table:1}
	\vspace{-3mm}
\end{table*}

\subsection{Ablation Study}\label{exp}
We perform ablation studies on the COCO dataset. It is widely used in general object detection and instance segmentation. In the ablation studies, ResNet-50~\cite{resnet} is used as the backbone network, and the same hyper-parameters as above are used. Specifically, our network is trained with stochastic gradient descent (SGD) for 1$\times$ training schedule~(\ie 12 epochs) with the initial learning rate being 0.01 and a mini-batch of 4 images per GPU. The learning rate is reduced by a factor of 10 at iteration 8 and 11 epochs, respectively. Weight decay and momentum are set as 0.0001 and 0.9, respectively.
We initialize our backbone networks with the weights pre-trained on ImageNet~\cite{imagenet}. The input images are resized to 768$\times$1280.

\subsubsection{Verification of Performance Upper Bound} 
The concern of polar representation is that it might not model the mask precisely. In this section, we show that this concern may not be necessary. Here we verify the upper bound of PolarMask as the IoU of the predicted mask and the ground-truth mask.
The verification results on different numbers of rays are shown in Figure~\ref{fig:upper bound}. 

We have the following observations:
(1) It can be seen that IoU is almost perfect~(\ie above 90\%) when the number of rays increases,  showing that polar representation is able to model the mask very well.
(2) However, when the rays are more than 72, the performance of polar representation encounters a bottleneck. For instance, 90 rays improve 0.4\% compare to 72 rays, and with 120 rays, the upper bound is saturated.
(3) It is more reasonable to use mass-center than bounding box-center as the center of an instance because the bounding box center is more likely to fall out of the instance.
From these observations, we can easily conclude that the concern about the upper bound of PolarMask++ is not necessary, and using a mass-center is better than a box-center.

\begin{figure*}[t]
\begin{center}
\includegraphics[width=0.88\textwidth]{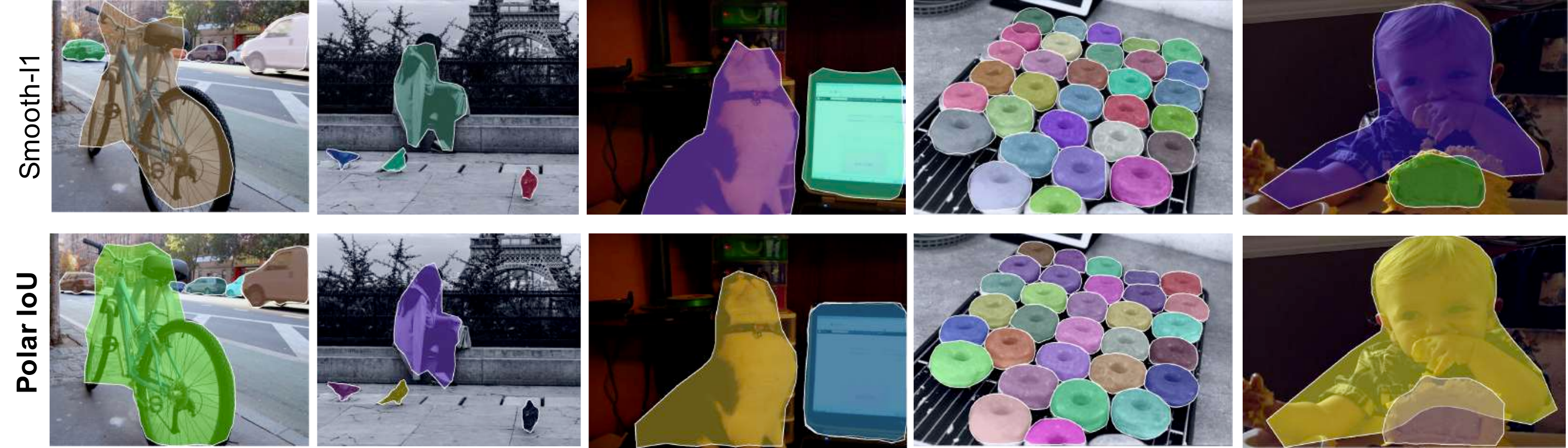}
\caption{Comparisons of visualization results of PolarMask++ when using Smooth-\L1 loss and Polar IoU loss. Polar IoU Loss achieves more accurate contour of instance, while Smooth-\L1 Loss exhibits artifacts.}
\label{fig:giraffe}
\end{center}
\vspace{-5mm}
\end{figure*}

\subsubsection{Number of Rays} 
\begin{figure}[t]
\begin{center}
\includegraphics[width=0.43\textwidth]{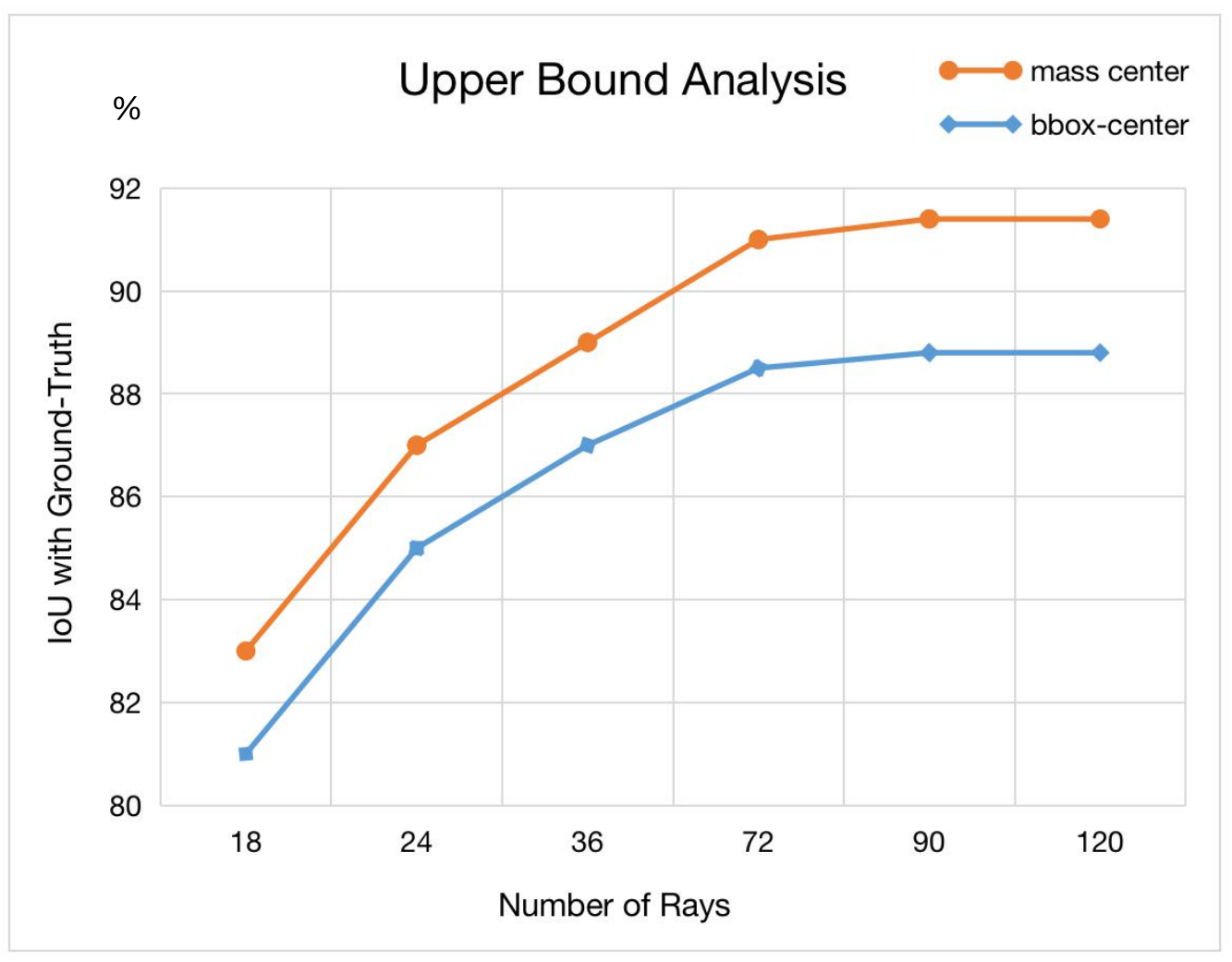}
\caption{Upper Bound Analysis. Larger number of rays would model instance masks with higher IoU. And mass-center is more effective to represent an instance than the box-center. For example, 90 rays improve 0.4\% compared to 72 rays, and the result is saturated when the number of rays approaches 120.}
\label{fig:upper bound}
\end{center}
\vspace{-5mm}
\end{figure}

The number of rays plays an important role in the entire system of PolarMask++. 
From Table~\ref{table:vectors} and Figure~\ref{fig:upper bound}, more rays achieve a higher upper bound and better AP. 
First, 24 rays improve 1.1\% AP compared to 18 rays, while 36 rays further improve 0.3\% AP compared to 24 rays. 
Second, too many rays (\eg 72 rays) would saturate the performance. The AP of 72 rays is 27.6\%, which is 0.1\% lower than 36 rays. 
We have two intuitive explanations.
(1) From the upper bound Figure~\ref{fig:upper bound}, when rays increasing from 36 to 72, although the upper bound keeps improving, the improvement is less than increasing the number of rays from 18 to 36.
In theory, it shows that there is not much room for improvement when the number of rays is 72.
(2) From the perspective of CNN, more rays indicate the network needs to learn more information, impeding network training.
According to the above discussions, we use 36 rays for PolarMask++, since 36 rays already depict the mask contours very well.

\subsubsection{Polar IoU Loss \emph{vs}. Smooth-\L1 Loss} 
We examine both Polar IoU Loss and Smooth-\L1 Loss in our architecture. We note that the regression loss of Smooth-\L1 Loss is \textit{significantly} larger than the classification loss since our architecture is a task of dense distance prediction. To cope with the imbalance, we select different factor $\alpha$ for the regression of Smooth-\L1 Loss. Experimental results are shown in Table~\ref{table:loss}. Our Polar IoU Loss achieves 27.7\%  AP without balancing regression loss and classification loss. In contrast, the best setting for Smooth-\L1 Loss achieves 25.1\%  AP (\ie 2.6\% worse than ours), showing that Polar IoU Loss is more effective than Smooth-\L1 loss for training the regression task of distances between mass-center and contours. 
The gap may come from two aspects. First, the Smooth-\L1 Loss may need more hyper-parameter search to achieve better performance, which can be time-consuming compared to the Polar IoU Loss. 
Second, Polar IoU Loss predicts all rays of one instance as a whole, which is superior to Smooth-\L1 Loss.

In Figure~\ref{fig:giraffe}, we also compare the results using the  Smooth-\L1 Loss and Polar IoU Loss respectively. Smooth-\L1 Loss exhibits systematic artifacts, suggesting that it lacks supervision of  the entire object. Polar IoU Loss shows more smooth and precise contours.

\subsubsection{Centerness Strategy}

In PolarMask~\cite{polarmask}, the Polar CenterNess is proposed to re-weight positive samples. The comparisons are shown in Table~\ref{table:centerness}. Polar Centerness improves by 1.4\% AP overall. 
Particularly, AP$_{75} $ and AP$_{L}$ considerably increase by 2.3\% and 2.6\% respectively. 
The reasons can be summarized as follows: (1) Our Polar CenterNess suppresses the scores of low-quality masks, and thus improve high-IoU metric (\ie AP75);
(2) In the original centerness, larger instances often have larger differences between maximum and minimum lengths of rays,
which is exactly the problem that polar centerness solves. 

In PolarMask++, we further improve polar centerness~\cite{polarmask} by proposing its soft version. 
From Table~\ref{table:soft}, we can find that on the one hand, soft polar centerness can improve more than 0.6\% AP compared to the original polar centerness. On the other hand, these three functions lead to nearly the same performance, which indicates that dividing rays of four subsets are essential to improve performance.

We explain the importance of dividing rays of four subsets as follows.
According to our observations, as the complex shapes of objects in COCO dataset, the value of original Polar Centerness tend to be low, because the $\min(\{d_{1}, d_{2}, \ldots, d_{n}\}$ are extremely imbalance with respect to the $\max(\{d_{1}, d_{2}, \ldots, d_{n}\}$ when object shapes are complicated.
As a result, the final classification scores tend to be low, which is harmful to the performance.
We hypothesize that the original polar centerness is too ``aggressive'', so that we propose a soft mechanism. The rays are divided into 4 subsets and we calculate centerness according to the value of these 4 subsets, which will ease the imbalance length problem of the original polar centerness.

\subsubsection{Strategy of Feature Pyramid}

The feature pyramid network (FPN) is another key component in PolarMask~\cite{polarmask}.
As show in Table~\ref{table:rfp}, the original FPN can achieve 29.8\% AP, while using PANet~\cite{panet} can only improve 0.1\% AP. 
In contrast, our Refined Feature Pyramid (RFP) can boost up the mAP by 0.4\%, especially  the performance of small and medium objects. 
It is reasonable since small objects are usually captured in the shallow layers of FPN, while large objects are usually targeted in the deeper layers. 
However, shallow layers have high resolution, lacking rich semantic information and relations between pixels. We use RFP to ease these two problems. 
First, the high-level and low-level features are aggregated and distributed again to help the information flow from deep layers to shallow layers. 
Second, the non-local module would help build the relationship of pixels on the feature maps.
As a result, the performance of small objects is significantly improved.

\textbf{Flops and params of FPN, PAN and RFP.}
To verify the efficiency of RFP compared with FPN, we calculate the flops and parameters of PolarMask++ with different necks. 
We set the input size to $1200 \times  800$. 
As shown in Table~\ref{table:rfp}, the flops of FPN/RFP/PAN is 252.5G/253.6G/258.5G, our RFP only adds 1.1G compared with FPN. However, the PAN's Flops is 258.5G, adding 3.0G Flops than FPN.
The parameters of FPN/RFP/PAN is 34.4/36.8/34.7M, our RFP only adds 0.3M parameters than FPN, the ratio of additional parameters is less than 1\%. 
However, PAN adds 2.4M parameters, adding 7\% parameters. But the PAN only improves 0.1 AP, while our RFP improves 0.4 AP.
In summary, the proposed RFP introduces only one parameterized operator, a non-local module, so the total flops and parameters improve minor. 
Moreover, the performance of RFP is also better than PAN, which typically needs more parameters and flops. It indicates the design of RFP is superior to PAN.

\subsubsection{Box Branch} 

Most of the previous methods of instance segmentation require localizing the bounding box of the object area and then segment the pixels inside the bounding box. For Instance, both Mask R-CNN and Yolact rely on box detection. In contrast, \textit{PolarMask++ is capable to directly produce the mask without bounding box detection}. 

In this section, we examine whether the additional bounding box detection branch would improve the mask AP. 
From Table~\ref{table:box}, we see that the bounding box branch contributes little to the performance of mask prediction. 
As mentioned before, the bounding box can be viewed as the simplest version of the mask with 4 rays only in polar representation.
Therefore, the regression of bounding boxes and masks are essentially similar tasks in PolarMask++, so that the improvement brought by multi-task training is limited. Moreover, unlike methods based on ``detect then segment'' paradigm, our framework does not need to detect the bounding box. 
We could use the predicted boxes to perform NMS or use the predicted masks to generate the minimum boxes to perform NMS.
In summary, box detection is not necessary for PolarMask++.
Thus, we do not have the bounding box prediction head in PolarMask++ for simplicity and faster speed.

\subsubsection{Backbone Architecture}
Table~\ref{table:backbone} shows the results of PolarMask++ when using different backbones. It can be seen that better features extracted by deeper and advanced  networks improve the performance. For example, PolarMask++ with ResNet-101 as backbone improves 1.4\% mAP compared to ResNet-50, while ResNeXt further improves 2.3\%.

\subsubsection{Speed \emph{vs}. Accuracy} 
Larger image sizes yield higher accuracy, but slower inference speeds. Table~\ref{table:speed} shows the speed and accuracy trade-off for different input image scales, which is represented by the shorter image side. 
The FPS is evaluated on one V100 GPU. Note that here we report the entire inference time, including all components. 
It shows that PolarMask++ has a strong potential as a real-time instance segmentation system with little modification.

\begin{table*}[t]
    \centering
    \scalebox{0.9}{
    \begin{tabular}{|c||c||c|c|c|c|c||c|c|c|c|c||c|}
\hline
\multicolumn{1}{|l||}{Refine} & AP   & AP$_{person}$ & AP$_{bicycle}$ & AP$_{dog}$ & AP$_{zebra}$& AP$_{giraffe}$ & AP$_{backpack}$ & AP$_{ball}$ & AP$_{cup}$ & AP$_{apple}$ & AP$_{clock}$ & Time(ms) \\ \hline
$\circ$                                & 30.2 & 37.8      & 12.5       & 45.1   & 42.3     & 33.2       & 14.9        & 38.8    & 38.3   & 18.2     & 48.4     & 0        \\ \hline
\checkmark                                & 34.1 & 43.2      & 14.2       & 55.2   & 53.1     & 51.0       & 14.6        & 38.1    & 38.5   & 18.1     & 47.9     & 1.5      \\ \hline
\end{tabular}}
    \vspace{2mm}
	\caption{Effectiveness of Mask Refinement. We show mAP and AP of different categories. In the last column, we report the time consumption for each object. We find that mask refinement largely improves the overall mAP from 30.2 to 34.1, improving 3.9 mAP. For specifical categories, on the one hand, some categories with complex and irregular shapes, \textit{e.g.} person and animals, their mAP is largely boosted up by mask refinement. On the other hand, other categories which has simple and regular shapes, \textit{e.g.} ball and clock, their mAP is hardly affected by mask refinement. The time consumption of mask refinement is minor, only 1.5ms per object.}
	\label{table:refine}
\end{table*}

\subsection{Mask Refinement with Post-processing FCN}
PolarMask++ performs well on scene text detection and cell instance segmentation, since the text and cell usually have regular shapes. However, it is challenging for PolarMask++ to get a very fine segmentation mask for an object with an irregular shape, \textit{e.g.} person and chair. To solve this problem, we straightforwardly add a post-processing FCN network to refine the predicted mask from PolarMask++. 
Implementation details and qualitative results are shown in appendix.

\textbf{Quantitative results.}
The results are shown in Table~\ref{table:refine}. The performance is largely improved from 30.2 AP to 34.1 AP, showing that the refinement FCN is useful.

On the one hand, we find that for some categories, \textit{e.g.} person, bicycle and animals, the original mask AP of PolarMask++ is relatively low, because they tend to have irregular shapes. 
However, when adding an additional refinement network, the mask AP largely improved. This result indicates that pixel-wise segmentation has advantages on objects with irregular shapes.

On the other hand, for the categories with regular shapes, \textit{e.g.} backpack, ball, cup, apple and clock, adding mask refinement network can not improve the mask AP, even a little drop. This result shows that polar representation has more advantages to segment objects with regular shapes.

The time consumption of refinement FCN is minor, only 1.5ms per object. So the mask refinement post-procedure is acceptable in practice.

\section{Conclusion}\label{sec:conclusion}
In this paper, we propose PolarMask++, which is a single shot anchor-box free method that unifies instance segmentation and rotated object detection. 
Different from previous works that typically solve mask prediction as binary classification in a spatial layout, PolarMask++ steps forward to represent a mask by its contour and model the contour by one center and rays emitted from the center to the contour in the polar coordinate space. PolarMask++ is designed almost as simple and clean as single-shot object detectors, introducing negligible computing overhead. We hope that the proposed PolarMask++ framework could provide a new perspective for single-shot instance segmentation and rotated object detection.
In the future, we would like to improve the capability of polar representation on objects with more complex shapes, which is the main factor to boost up performance on the COCO dataset.

\ifCLASSOPTIONcompsoc

\section*{Acknowledgments}
\else
  \section*{Acknowledgment}
\fi

This work was partially supported by the RGC General Research Fund of HK No.27208720, HKU Seed Fund for Basic Research, Start-up Fund and Research Donation from SenseTime.

\ifCLASSOPTIONcaptionsoff
  \newpage
\fi

{\small
\bibliographystyle{ieee}
\bibliography{reference}

\begin{thebibliography}{10}\itemsep=-1pt

\bibitem{sapr}
Y.~Bi and Z.~Hu.
\newblock Scale-aware polar representation for arbitrarily-shaped text
  detection.
\newblock In {\em Proceedings of the Asian Conference on Computer Vision},
  2020.

\bibitem{yolact}
D.~Bolya, C.~Zhou, F.~Xiao, and Y.~J. Lee.
\newblock Yolact: {Real-time} instance segmentation.
\newblock {\em Proc. IEEE Int. Conf. Comp. Vis.}, 2019.

\bibitem{dcan}
H.~Chen, X.~Qi, L.~Yu, and P.-A. Heng.
\newblock Dcan: deep contour-aware networks for accurate gland segmentation.
\newblock In {\em Proceedings of the IEEE conference on Computer Vision and
  Pattern Recognition}, pages 2487--2496, 2016.

\bibitem{mmdetection}
K.~Chen, J.~Wang, J.~Pang, Y.~Cao, Y.~Xiong, X.~Li, S.~Sun, W.~Feng, Z.~Liu,
  J.~Xu, Z.~Zhang, D.~Cheng, C.~Zhu, T.~Cheng, Q.~Zhao, B.~Li, X.~Lu, R.~Zhu,
  Y.~Wu, J.~Dai, J.~Wang, J.~Shi, W.~Ouyang, C.~C. Loy, and D.~Lin.
\newblock Mmdetection: Open mmlab detection toolbox and benchmark, 2019.

\bibitem{tensormask}
X.~Chen, R.~Girshick, K.~He, and P.~Doll{\'a}r.
\newblock Tensormask: A foundation for dense object segmentation.
\newblock {\em Proc. IEEE Int. Conf. Comp. Vis.}, 2019.

\bibitem{darnet}
D.~Cheng, R.~Liao, S.~Fidler, and R.~Urtasun.
\newblock Darnet: Deep active ray network for building segmentation.
\newblock In {\em Proceedings of the IEEE Conference on Computer Vision and
  Pattern Recognition}, pages 7431--7439, 2019.

\bibitem{mnc}
J.~Dai, K.~He, and J.~Sun.
\newblock Instance-aware semantic segmentation via multi-task network cascades.
\newblock In {\em Proc. IEEE Conf. Comp. Vis. Patt. Recogn.}, pages 3150--3158,
  2016.

\bibitem{rfcn}
J.~Dai, Y.~Li, K.~He, and J.~Sun.
\newblock R-fcn: Object detection via region-based fully convolutional
  networks.
\newblock In {\em Proc. Advances in Neural Inf. Process. Syst.}, pages
  379--387, 2016.

\bibitem{dcn}
J.~Dai, H.~Qi, Y.~Xiong, Y.~Li, G.~Zhang, H.~Hu, and Y.~Wei.
\newblock Deformable convolutional networks.
\newblock In {\em Proc. IEEE Int. Conf. Comp. Vis.}, pages 764--773, 2017.

\bibitem{PixelLink}
D.~Deng, H.~Liu, X.~Li, and D.~Cai.
\newblock Pixellink: Detecting scene text via instance segmentation.
\newblock In {\em Proc. {AAAI} Conf. Artificial Intell.}, 2018.

\bibitem{imagenet}
J.~Deng, W.~Dong, R.~Socher, L.-J. Li, K.~Li, and L.~Fei-Fei.
\newblock Imagenet: A large-scale hierarchical image database.
\newblock In {\em Proc. IEEE Conf. Comp. Vis. Patt. Recogn.}, pages 248--255.
  Ieee, 2009.

\bibitem{activeray}
J.~Denzler and H.~Niemann.
\newblock Active rays: Polar-transformed active contours for real-time contour
  tracking.
\newblock {\em Real-Time Imaging}, 5(3):203--213, 1999.

\bibitem{centernet}
K.~Duan, S.~Bai, L.~Xie, H.~Qi, Q.~Huang, and Q.~Tian.
\newblock Centernet: Keypoint triplets for object detection.
\newblock In {\em Proc. IEEE Int. Conf. Comp. Vis.}, pages 6569--6578, 2019.

\bibitem{textdragon}
W.~Feng, W.~He, F.~Yin, X.-Y. Zhang, and C.-L. Liu.
\newblock Textdragon: An end-to-end framework for arbitrary shaped text
  spotting.
\newblock In {\em Proceedings of the IEEE International Conference on Computer
  Vision}, 2019.

\bibitem{dssd}
C.-Y. Fu, W.~Liu, A.~Ranga, A.~Tyagi, and A.~C. Berg.
\newblock Dssd: Deconvolutional single shot detector.
\newblock {\em arXiv preprint arXiv:1701.06659}, 2017.

\bibitem{fastrcnn}
R.~Girshick.
\newblock Fast {R-CNN}.
\newblock In {\em Proc. IEEE Int. Conf. Comp. Vis.}, pages 1440--1448, 2015.

\bibitem{rcnn}
R.~Girshick, J.~Donahue, T.~Darrell, and J.~Malik.
\newblock Rich feature hierarchies for accurate object detection and semantic
  segmentation.
\newblock In {\em Proc. IEEE Conf. Comp. Vis. Patt. Recogn.}, pages 580--587,
  2014.

\bibitem{Detectron}
R.~Girshick, I.~Radosavovic, G.~Gkioxari, P.~Doll\'{a}r, and K.~He.
\newblock Detectron.
\newblock \url{https://github.com/facebookresearch/detectron}, 2018.

\bibitem{dsb}
B.~A. Hamilton.
\newblock Kaggle. 2018 data science bowl: Find the nuclei in divergent images
  to advance medical discovery, 2018.

\bibitem{maskrcnn}
K.~He, G.~Gkioxari, P.~Doll{\'a}r, and R.~Girshick.
\newblock {Mask R-CNN}.
\newblock In {\em Proc. IEEE Int. Conf. Comp. Vis.}, pages 2961--2969, 2017.

\bibitem{sppnet}
K.~He, X.~Zhang, S.~Ren, and J.~Sun.
\newblock Spatial pyramid pooling in deep convolutional networks for visual
  recognition.
\newblock {\em IEEE transactions on pattern analysis and machine intelligence},
  37(9):1904--1916, 2015.

\bibitem{resnet}
K.~He, X.~Zhang, S.~Ren, and J.~Sun.
\newblock Deep residual learning for image recognition.
\newblock In {\em Proc. IEEE Conf. Comp. Vis. Patt. Recogn.}, June 2016.

\bibitem{he2017single}
P.~He, W.~Huang, T.~He, Q.~Zhu, Y.~Qiao, and X.~Li.
\newblock Single shot text detector with regional attention.
\newblock In {\em Proc. IEEE Int. Conf. Comp. Vis.}, 2017.

\bibitem{deepreg}
W.~He, X.-Y. Zhang, F.~Yin, and C.-L. Liu.
\newblock Deep direct regression for multi-oriented scene text detection.
\newblock In {\em Proc. IEEE Int. Conf. Comp. Vis.}, 2017.

\bibitem{patchperpix}
P.~Hirsch, L.~Mais, and D.~Kainmueller.
\newblock Patchperpix for instance segmentation.
\newblock {\em arXiv preprint arXiv:2001.07626}, 2020.

\bibitem{hu2017wordsup}
H.~Hu, C.~Zhang, Y.~Luo, Y.~Wang, J.~Han, and E.~Ding.
\newblock Wordsup: Exploiting word annotations for character based text
  detection.
\newblock In {\em Proceedings of the IEEE International Conference on Computer
  Vision}, pages 4940--4949, 2017.

\bibitem{densebox}
L.~Huang, Y.~Yang, Y.~Deng, and Y.~Yu.
\newblock Densebox: Unifying landmark localization with end to end object
  detection.
\newblock {\em arXiv preprint arXiv:1509.04874}, 2015.

\bibitem{msrcnn}
Z.~Huang, L.~Huang, Y.~Gong, C.~Huang, and X.~Wang.
\newblock Mask scoring r-cnn.
\newblock In {\em Proc. IEEE Conf. Comp. Vis. Patt. Recogn.}, pages 6409--6418,
  2019.

\bibitem{2015icdar}
D.~Karatzas, L.~Gomez{-}Bigorda, A.~Nicolaou, S.~K. Ghosh, A.~D. Bagdanov,
  M.~Iwamura, J.~Matas, L.~Neumann, V.~R. Chandrasekhar, S.~Lu, F.~Shafait,
  S.~Uchida, and E.~Valveny.
\newblock {ICDAR} 2015 competition on robust reading.
\newblock In {\em Proc. ICDAR}, pages 1156--1160, 2015.

\bibitem{foveabox}
T.~Kong, F.~Sun, H.~Liu, Y.~Jiang, and J.~Shi.
\newblock Foveabox: Beyond anchor-based object detector.
\newblock {\em arXiv preprint arXiv:1904.03797}, 2019.

\bibitem{hypernet}
T.~Kong, A.~Yao, Y.~Chen, and F.~Sun.
\newblock Hypernet: Towards accurate region proposal generation and joint
  object detection.
\newblock In {\em Proceedings of the IEEE conference on computer vision and
  pattern recognition}, pages 845--853, 2016.

\bibitem{cornernet}
H.~Law and J.~Deng.
\newblock Cornernet: Detecting objects as paired keypoints.
\newblock In {\em Proceedings of the European Conference on Computer Vision
  (ECCV)}, pages 734--750, 2018.

\bibitem{fcis}
Y.~Li, H.~Qi, J.~Dai, X.~Ji, and Y.~Wei.
\newblock Fully convolutional instance-aware semantic segmentation.
\newblock In {\em Proc. IEEE Conf. Comp. Vis. Patt. Recogn.}, pages 2359--2367,
  2017.

\bibitem{textboxes}
M.~Liao, B.~Shi, X.~Bai, X.~Wang, and W.~Liu.
\newblock Textboxes: A fast text detector with a single deep neural network.
\newblock In {\em Thirty-First AAAI Conference on Artificial Intelligence},
  2017.

\bibitem{rrd}
M.~Liao, Z.~Zhu, B.~Shi, G.-s. Xia, and X.~Bai.
\newblock Rotation-sensitive regression for oriented scene text detection.
\newblock In {\em Proceedings of the IEEE Conference on Computer Vision and
  Pattern Recognition}, pages 5909--5918, 2018.

\bibitem{fpn}
T.-Y. Lin, P.~Dollar, R.~Girshick, K.~He, B.~Hariharan, and S.~Belongie.
\newblock Feature pyramid networks for object detection.
\newblock In {\em Proc. IEEE Conf. Comp. Vis. Patt. Recogn.}, July 2017.

\bibitem{focalloss}
T.-Y. Lin, P.~Goyal, R.~Girshick, K.~He, and P.~Dollar.
\newblock Focal loss for dense object detection.
\newblock In {\em Proc. IEEE Int. Conf. Comp. Vis.}, Oct 2017.

\bibitem{coco}
T.-Y. Lin, M.~Maire, S.~Belongie, J.~Hays, P.~Perona, D.~Ramanan,
  P.~Doll{\'a}r, and C.~L. Zitnick.
\newblock Microsoft coco: Common objects in context.
\newblock In {\em Proc. Eur. Conf. Comp. Vis.}, pages 740--755. Springer, 2014.

\bibitem{panet}
S.~Liu, L.~Qi, H.~Qin, J.~Shi, and J.~Jia.
\newblock Path aggregation network for instance segmentation.
\newblock In {\em Proc. IEEE Conf. Comp. Vis. Patt. Recogn.}, pages 8759--8768,
  2018.

\bibitem{SSD}
W.~Liu, D.~Anguelov, D.~Erhan, C.~Szegedy, S.~Reed, C.-Y. Fu, and A.~C. Berg.
\newblock {SSD}: Single shot multibox detector.
\newblock In {\em Proc. Eur. Conf. Comp. Vis.}, October 2016.

\bibitem{mcn}
Z.~Liu, G.~Lin, S.~Yang, J.~Feng, W.~Lin, and W.~L. Goh.
\newblock Learning markov clustering networks for scene text detection.
\newblock {\em Proc. IEEE Conf. Comp. Vis. Patt. Recogn.}, 2018.

\bibitem{fcn}
J.~Long, E.~Shelhamer, and T.~Darrell.
\newblock Fully convolutional networks for semantic segmentation.
\newblock In {\em Proceedings of the IEEE conference on computer vision and
  pattern recognition}, pages 3431--3440, 2015.

\bibitem{textsnake}
S.~Long, J.~Ruan, W.~Zhang, X.~He, W.~Wu, and C.~Yao.
\newblock Textsnake: A flexible representation for detecting text of arbitrary
  shapes.
\newblock {\em Proc. Eur. Conf. Comp. Vis.}, 2018.

\bibitem{nrcnn}
G.~Lv, K.~Wen, Z.~Wu, X.~Jin, H.~An, and J.~He.
\newblock Nuclei r-cnn: Improve mask r-cnn for nuclei segmentation.
\newblock In {\em 2019 IEEE 2nd International Conference on Information
  Communication and Signal Processing (ICICSP)}, pages 357--362. IEEE, 2019.

\bibitem{lyu2018multi}
P.~Lyu, C.~Yao, W.~Wu, S.~Yan, and X.~Bai.
\newblock Multi-oriented scene text detection via corner localization and
  region segmentation.
\newblock {\em arXiv preprint arXiv:1802.08948}, 2018.

\bibitem{rrpn}
J.~Ma, W.~Shao, H.~Ye, L.~Wang, H.~Wang, Y.~Zheng, and X.~Xue.
\newblock Arbitrary-oriented scene text detection via rotation proposals.
\newblock {\em IEEE Transactions on Multimedia}, 20(11):3111--3122, 2018.

\bibitem{pang2019libra}
J.~Pang, K.~Chen, J.~Shi, H.~Feng, W.~Ouyang, and D.~Lin.
\newblock Libra r-cnn: Towards balanced learning for object detection.
\newblock In {\em Proceedings of the IEEE Conference on Computer Vision and
  Pattern Recognition}, pages 821--830, 2019.

\bibitem{deepsnake}
S.~Peng, W.~Jiang, H.~Pi, X.~Li, H.~Bao, and X.~Zhou.
\newblock Deep snake for real-time instance segmentation.
\newblock In {\em Proceedings of the IEEE/CVF Conference on Computer Vision and
  Pattern Recognition}, pages 8533--8542, 2020.

\bibitem{yolo}
J.~Redmon, S.~Divvala, R.~Girshick, and A.~Farhadi.
\newblock You only look once: Unified, real-time object detection.
\newblock In {\em Proc. IEEE Conf. Comp. Vis. Patt. Recogn.}, pages 779--788,
  2016.

\bibitem{yolov2}
J.~Redmon and A.~Farhadi.
\newblock Yolo9000: Better, faster, stronger.
\newblock In {\em Proc. IEEE Conf. Comp. Vis. Patt. Recogn.}, July 2017.

\bibitem{fasterrcnn}
S.~Ren, K.~He, R.~Girshick, and J.~Sun.
\newblock Faster r-cnn: Towards real-time object detection with region proposal
  networks.
\newblock In {\em Proc. Advances in Neural Inf. Process. Syst.}, pages 91--99,
  2015.

\bibitem{unet}
O.~Ronneberger, P.~Fischer, and T.~Brox.
\newblock U-net: Convolutional networks for biomedical image segmentation.
\newblock In {\em International Conference on Medical image computing and
  computer-assisted intervention}, pages 234--241. Springer, 2015.

\bibitem{celldet}
U.~Schmidt, M.~Weigert, C.~Broaddus, and G.~Myers.
\newblock Cell detection with star-convex polygons.
\newblock In {\em Proc.\ Int.\ Medical Image Computing and Computer-Assisted
  Intervention}, pages 265--273. Springer, 2018.

\bibitem{shi2017detecting}
B.~Shi, X.~Bai, and S.~Belongie.
\newblock Detecting oriented text in natural images by linking segments.
\newblock In {\em Proc. IEEE Conf. Comp. Vis. Patt. Recogn.}, 2017.

\bibitem{tian2016detecting}
Z.~Tian, W.~Huang, T.~He, P.~He, and Y.~Qiao.
\newblock Detecting text in natural image with connectionist text proposal
  network.
\newblock In {\em Proc. Eur. Conf. Comp. Vis.}, 2016.

\bibitem{FCOS}
Z.~Tian, C.~Shen, H.~Chen, and T.~He.
\newblock {FCOS}: Fully convolutional one-stage object detection.
\newblock In {\em Proc. IEEE Int. Conf. Comp. Vis.}, 2019.

\bibitem{textray}
F.~Wang, Y.~Chen, F.~Wu, and X.~Li.
\newblock Textray: Contour-based geometric modeling for arbitrary-shaped scene
  text detection.
\newblock In {\em Proceedings of the 28th ACM International Conference on
  Multimedia}, pages 111--119, 2020.

\bibitem{psenet}
W.~Wang, E.~Xie, X.~Li, W.~Hou, T.~Lu, G.~Yu, and S.~Shao.
\newblock Shape robust text detection with progressive scale expansion network.
\newblock In {\em Proceedings of the IEEE Conference on Computer Vision and
  Pattern Recognition}, pages 9336--9345, 2019.

\bibitem{pan}
W.~Wang, E.~Xie, X.~Song, Y.~Zang, W.~Wang, T.~Lu, G.~Yu, and C.~Shen.
\newblock Efficient and accurate arbitrary-shaped text detection with pixel
  aggregation network.
\newblock In {\em Proceedings of the IEEE International Conference on Computer
  Vision}, pages 8440--8449, 2019.

\bibitem{nonlocal}
X.~Wang, R.~Girshick, A.~Gupta, and K.~He.
\newblock Non-local neural networks.
\newblock In {\em Proceedings of the IEEE conference on computer vision and
  pattern recognition}, pages 7794--7803, 2018.

\bibitem{polarmask}
E.~Xie, P.~Sun, X.~Song, W.~Wang, X.~Liu, D.~Liang, C.~Shen, and P.~Luo.
\newblock Polarmask: Single shot instance segmentation with polar
  representation.
\newblock {\em arXiv preprint arXiv:1909.13226}, 2019.

\bibitem{spcnet}
E.~Xie, Y.~Zang, S.~Shao, G.~Yu, C.~Yao, and G.~Li.
\newblock Scene text detection with supervised pyramid context network.
\newblock In {\em Proceedings of the AAAI Conference on Artificial
  Intelligence}, volume~33, pages 9038--9045, 2019.

\bibitem{ese_seg}
W.~Xu, H.~Wang, F.~Qi, and C.~Lu.
\newblock Explicit shape encoding for real-time instance segmentation.
\newblock In {\em Proc. IEEE Int. Conf. Comp. Vis.}, pages 5168--5177, 2019.

\bibitem{textfield}
Y.~Xu, Y.~Wang, W.~Zhou, Y.~Wang, Z.~Yang, and X.~Bai.
\newblock Textfield: learning a deep direction field for irregular scene text
  detection.
\newblock {\em IEEE Transactions on Image Processing}, 28(11):5566--5579, 2019.

\bibitem{msr}
C.~Xue, S.~Lu, and W.~Zhang.
\newblock Msr: multi-scale shape regression for scene text detection.
\newblock {\em arXiv preprint arXiv:1901.02596}, 2019.

\bibitem{reppoints}
Z.~Yang, S.~Liu, H.~Hu, L.~Wang, and S.~Lin.
\newblock Reppoints: Point set representation for object detection.
\newblock {\em arXiv: Comp. Res. Repository}, 2019.

\bibitem{msra}
C.~Yao, X.~Bai, W.~Liu, Y.~Ma, and Z.~Tu.
\newblock Detecting texts of arbitrary orientations in natural images.
\newblock In {\em 2012 IEEE conference on computer vision and pattern
  recognition}, pages 1083--1090. IEEE, 2012.

\bibitem{yi2019object}
J.~Yi, H.~Tang, P.~Wu, B.~Liu, D.~J. Hoeppner, D.~N. Metaxas, L.~Han, and
  W.~Fan.
\newblock Object-guided instance segmentation for biological images.
\newblock {\em arXiv preprint arXiv:1911.09199}, 2019.

\bibitem{kpgraph}
J.~Yi, P.~Wu, Q.~Huang, H.~Qu, B.~Liu, D.~J. Hoeppner, and D.~N. Metaxas.
\newblock Multi-scale cell instance segmentation with keypoint graph based
  bounding boxes.
\newblock In {\em International Conference on Medical Image Computing and
  Computer-Assisted Intervention}, pages 369--377. Springer, 2019.

\bibitem{unitbox}
J.~Yu, Y.~Jiang, Z.~Wang, Z.~Cao, and T.~Huang.
\newblock Unitbox: An advanced object detection network.
\newblock In {\em Proc.\ ACM Int.\ Conf.\ Multimedia}, pages 516--520. ACM,
  2016.

\bibitem{polardet}
P.~Zhao, Z.~Qu, Y.~Bu, W.~Tan, Y.~Ren, and S.~Pu.
\newblock Polardet: A fast, more precise detector for rotated target in aerial
  images.
\newblock {\em arXiv preprint arXiv:2010.08720}, 2020.

\bibitem{zhou2020objects}
L.~Zhou, H.~Wei, H.~Li, Y.~Zhang, X.~Sun, and W.~Zhao.
\newblock Objects detection for remote sensing images based on polar
  coordinates.
\newblock {\em arXiv preprint arXiv:2001.02988}, 2020.

\bibitem{objectspoints}
X.~Zhou, D.~Wang, and P.~Kr{\"a}henb{\"u}hl.
\newblock Objects as points.
\newblock {\em arXiv: Comp. Res. Repository}, 2019.

\bibitem{east}
X.~Zhou, C.~Yao, H.~Wen, Y.~Wang, S.~Zhou, W.~He, and J.~Liang.
\newblock East: an efficient and accurate scene text detector.
\newblock In {\em Proceedings of the IEEE conference on Computer Vision and
  Pattern Recognition}, pages 5551--5560, 2017.

\bibitem{extremenet}
X.~Zhou, J.~Zhuo, and P.~Krahenbuhl.
\newblock Bottom-up object detection by grouping extreme and center points.
\newblock In {\em Proc. IEEE Conf. Comp. Vis. Patt. Recogn.}, pages 850--859,
  2019.

\end{thebibliography}
}

\end{document}